\pdfoutput=1

\documentclass[11pt]{article}

\usepackage[]{ACL2023}

\usepackage{times}
\usepackage{latexsym}

\usepackage[T1]{fontenc}

\usepackage[utf8]{inputenc}

\usepackage{microtype}

\usepackage{inconsolata}

\usepackage{multirow}
\usepackage{placeins}
\usepackage{graphicx} 

\usepackage{adjustbox}

\title{Scalability of Bayesian Network Structure Elicitation with Large Language Models: a Novel Methodology and Comparative Analysis}

\author{Nikolay Babakov \\
  Universidade de \\ Santiago de Compostela \\
  \texttt{nikolay.babakov@usc.es} \\\And
  Ehud Reiter \\
  University of Aberdeen \\
  \texttt{e.reiter@abdn.ac.uk} \\\And
    Alberto Bugarín \\
  Universidade de \\ Santiago de Compostela \\
  \texttt{alberto.bugarin.diz@usc.es}}

\begin{document}
\maketitle
\begin{abstract}
In this work, we propose a novel method for Bayesian Networks (BNs) structure elicitation that is based on the initialization of several LLMs with different experiences, independently querying them to create a structure of the BN, and further obtaining the final structure by majority voting. We compare the method with one alternative method on various widely and not widely known BNs of different sizes and study the scalability of both methods on them. We also propose an approach to check the contamination of BNs in LLM, which shows that some widely known BNs are inapplicable for testing the LLM usage for BNs structure elicitation. We also show that some BNs may be inapplicable for such experiments because their node names are indistinguishable. The experiments on the other BNs show that our method performs better than the existing method with one of the three studied LLMs; however, the performance of both methods significantly decreases with the increase in BN size.
\end{abstract}

\section{Introduction}
\label{sec:intro}

\begin{figure}[]
    \centering
    \includegraphics[scale=0.75]{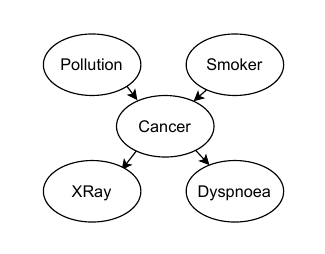}
    \caption{A BN for the lung cancer problem. Note that conditional probability tables encoding the probabilistic relations between variables are not shown here.}
    \label{fig:bn_sample}
\end{figure}

A Bayesian network (BN) is a probabilistic graphical model that represents a set of variables and their conditional dependencies through a directed acyclic graph (DAG). Refer to Figure~\ref{fig:bn_sample} for a simple BN~\cite{korb2010bayesian} which describes the toy scenario involving some possible reasons (Pollution and Smoker) and consequences (XRay results and Dyspnoea) of Lung Cancer.

The DAG encoding causal relationships among BN nodes is a crucial part of a BN. It can be learned from data using various BN structure learning (BNSL) algorithms~\cite{kitson2023survey} or from expert knowledge~\cite{https://doi.org/10.1111/risa.13759}. However, the recent progress of Large Language Models (LLMs) encouraged researchers to try using LLMs as another source of information to build a BN structure~\cite{ban2023query,long2023causal}.

In this paper, we propose a new approach of LLM-driven BNSL that takes inspiration from the real-world practices of collective experts' discussion of BN structure, in particular, the Delphi protocol~\cite{rowe1991delphi}. It assumes the presence of multiple experts who asynchronously and anonymously discuss a group of complex questions. We transfer this intuition into the LLM environment: we generate the expertise of several ``LLM experts'', initialize different instances of LLMs with this expertise, ask them similar questions about BN structure, and, in the end, define the finally elicited structure by majority vote. 

Even though LLMs show impressive performance on various casual tasks~\cite{tu2023causal}, it is still necessary to understand their limitations~\cite{tamkin2021understanding}. In terms of BNSL, the knowledge about certain BN can be obtained by an LLM during its training, which yields data contamination and results in an inadequate increase in the quality of the particular task~\cite{sainz2023nlp}. To the best of our knowledge, the existing works about LLM usage for BNSL run their experiments with widely-known BNs disregarding the risk of data contamination. We propose a simple prompting technique that gives a general idea of whether the information about a particular BN is known to the LLM and also involve not widely known BNs in the experiments. 

The experiments performed in this work involve data contamination check and LLM-driven BNSL with our original and the existing method~\cite{ban2023query} on the BNs of different sizes (from several to hundreds of nodes) and from different sources (widely known and frequently used in BNSL papers and less known obtained from the sources less likely available to LLMs training).

The code of the developed method of LLM-driven BNSL is publicly available for easiness of reproduction.\footnote{\url{https://gitlab.nl4xai.eu/nikolay.babakov/delphi_lm_xpertnet}}.



\section{Related works}
\label{sec:related}


The structure of BN is its crucial element. If the data for a particular task is available, the structure of BN can be learned from it using structure learning algorithms~\cite{kitson2023survey}. Otherwise, the structure can be obtained through a collective discussion between human experts. One of the optimal ways to perform such discussion is by using the Delphi protocol~\cite{rowe1991delphi,turoff2002delphi}, which engages multiple experts in discussing certain problems asynchronously and anonymously under the review of the facilitator. This technique was first used for the specific task of BN structure and parameters elicitation in~\cite{nicholson2016delphi}. Delphi protocol usage was further expanded to the whole pipeline of BN structure elicitation in the BARD system~\cite{https://doi.org/10.1111/risa.13759}. 



LLMs have shown impressive results in many tasks requiring causal analysis~\cite{nori2023capabilities,tu2023causal}. The recent best practices of LLM prompting can imitate the collective mindset similar to the Delphi-style discussion. First, the LLM could be associated with a particular role on behalf of which it is asked to answer the prompt~\cite{li2023camel}. Second, there already are numerous papers that engage different instances of LLMs with different ``experience'' ~\cite{li2023prd,wang2023unleashing,gao2023s} or even models of different modalities (e.g., visual and language)~\cite{zhuge2023mindstorms} into discussing a target task. 


However, while designing any LLM-driven method, it is necessary to take into account the phenomenon of data contamination~\cite{sainz2023nlp,sainz-etal-2023-nlp}. This problem is actively discussed~\cite{jacovi2023stop} and numerous approaches to data contamination testing have been proposed. Some of them rely on the log-likelihood of the LLM output~\cite{oren2023proving,shi2023detecting}, on the exact data that is suspected to be contaminated~\cite{elangovan2021memorization,golchin2023time} or on special tasks~\cite{mccoy2023embers} and quizzes~\cite{golchin2023data}.

\begin{table}[b!]
\centering
\begin{adjustbox}{width=\linewidth}
\begin{tabular}{|l|l|l|l|}
\hline
\textbf{BN name} & \textbf{Nodes} & \textbf{Edges} & \textbf{Source} \\ \hline
coma & 5 & 5 & \href{https://repo.bayesfusion.com/bayesbox.html}{bayesfusion} \\ \hline
covid & 20 & 26 & \href{https://repo.bayesfusion.com/bayesbox.html}{bayesfusion} \\ \hline
insurance & 27 & 52 & \href{https://www.bnlearn.com/bnrepository/}{bnlearn} \\ \hline
alarm & 37 & 46 & \href{https://www.bnlearn.com/bnrepository/}{bnlearn} \\ \hline
barley & 48 & 84 & \href{https://www.bnlearn.com/bnrepository/}{bnlearn} \\ \hline
hailfinder & 56 & 66& \href{https://www.bnlearn.com/bnrepository/}{bnlearn} \\ \hline
pathfinder & 109 & 195 & \href{https://www.bnlearn.com/bnrepository/}{bnlearn} \\ \hline
andes & 223 & 338 & \href{https://www.bnlearn.com/bnrepository/}{bnlearn} \\ \hline
diabetes & 413 & 602 & \href{https://www.bnlearn.com/bnrepository/}{bnlearn} \\ \hline
munin & 186 & 273 & \href{https://www.bnlearn.com/bnrepository/}{bnlearn} \\ \hline
hepar2 & 70 & 123 & \href{https://www.bnlearn.com/bnrepository/}{bnlearn} \\ \hline
agro & 6 & 10 & \cite{baudrit2022decision} \\ \hline
sperm & 9 & 11 & \cite{samie2022use} \\ \hline
screen & 16 & 21 & \cite{zio2022bayesian} \\ \hline
sids & 17 & 27 & \cite{hamayasu2022proof} \\ \hline
apple & 29 & 62 & \cite{sottocornola2023development} \\ \hline
urinary & 36 & 107 & \cite{ramsay2022urinary} \\ \hline
\end{tabular}
\end{adjustbox}
\caption{Bayesian Networks used for experiments.}
\label{tab:bn_size_source}
\end{table}

Naturally, most LLM applications are directed toward fields where the information could normally be described in plain text (e.g. education~\cite{KASNECI2023102274}). However, the representation of graph structures in plain text is not that straightforward, but some workarounds are possible. For example,~\cite{wang2023can} proposed specific prompting that allows using LLMs for solving general graph problems (e.g., connectivity, maximum flow). At the same time, several papers dedicated to the specific task of LLM-driven causal graph construction already exist. ~\cite{long2023can} showed that using even a very simple prompt may demonstrate proper performance of causal relations extraction. ~\cite{long2023causal} use LLMs as ``imperfect experts'' that construct BN's structure. \cite{vashishtha2023causal} proposes four types of prompts, the most effective of which is a prompting strategy that asks LLM to elicit causal relations between all possible triplets of BN nodes.~\cite{ban2023query} proposes to use three consecutive prompts to understand the meaning of nodes, construct causal relations, and check LLM's own reasoning.


\section{Experimental setup}
\label{sec:experiment_setup}

\subsection{Bayesian Networks}
\label{sec:engaged_bns}

We aim to use BNs of different sizes and likelihoods of being seen by LLMs during training. The ``popular'' BNs are obtained from bayesfusion\footnote{\url{https://repo.bayesfusion.com/bayesbox.html}} and bnlearn\footnote{\url{https://www.bnlearn.com/bnrepository/}} websites. The ``unpopular'' BNs are obtained from the papers dedicated to BN application to specific problems in various domains. See the list of all BNs engaged in our experiments in Table~\ref{tab:bn_size_source}. We succeeded in finding large BNs only from the bnlearn website, because in the case of large BNs developed in the research papers, even their structure is hard to fit into the paper, and authors rarely distribute their BNs in a standalone runnable file. 

\subsection{Method of BN structure learning using LLMs}

\subsubsection{New method for BN structure elicitation}

\begin{figure*}[ht!]
    \centering
    \includegraphics[scale=0.75]{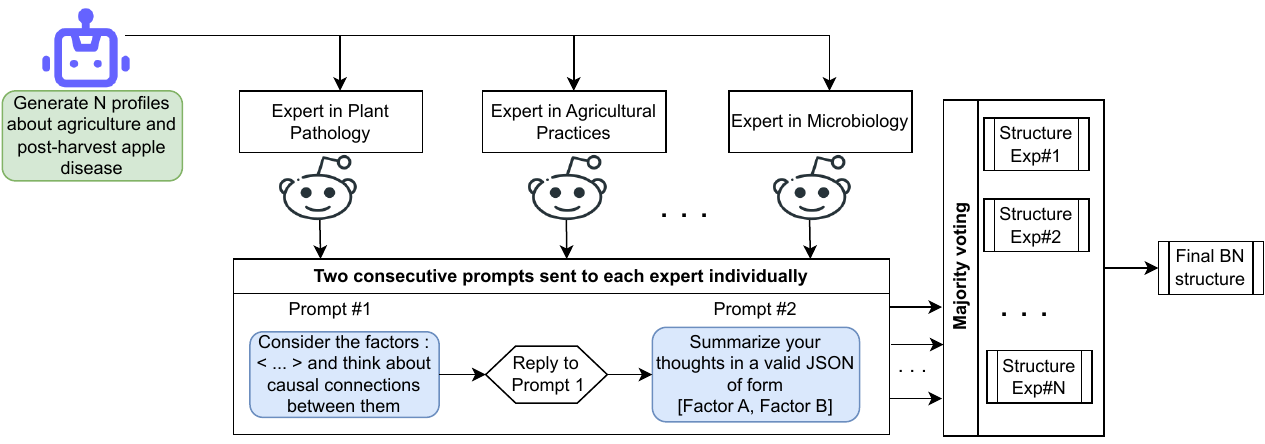}
    \caption{The overview of the proposed method for BN structure elicitation. Facilitator LLM generates N profiles relevant to the given BN. ``LLM experts'' initialized with different profiles are queried about the structure of BN using two consecutive prompts. The final BN structure is obtained by majority voting between the structures elicited from individual experts.}
    \label{fig:our_method_overview}
\end{figure*}


We propose a novel method for eliciting knowledge from LLM to construct a BN structure. In the proposed method, we imitate the Delphi-style discussion (experts discuss complex problems anonymously and asynchronously) for the construction of BN structure using LLMs' knowledge. Fig.~\ref{fig:our_method_overview} shows the general idea of our approach. We assume that some basic information about the BN is available: its aim and the area of necessary knowledge. First, we query facilitator LLM; its task is to generate an arbitrary odd number of experts who have unique experience relevant to the necessary knowledge area. Second, we initialize multiple LLM threads with their unique experiences and send them two consecutive prompts. The first prompt contains a list of node names and a request to discuss what kind of causal relations could be between these nodes (the node names are preliminarily clarified using \textit{Understand} prompt discussed in Section~\ref{sec:harness_intro}). The second prompt requests LLM to summarize the reasoning from the answer to the first prompt into a JSON structure. The final BN structure is obtained by majority voting - the edge is added to the structure only if more than half of the ``LLM experts'' voted for it. If the cycle is found in the structure by a certain expert, it is resolved with an additional request about the conflicting pair of nodes. The full prompts for all aforementioned steps are available in Appendix~\ref{sec:app_our_method}.








\subsubsection{Alternative method}
\label{sec:harness_intro}


We aim to compare our method with an alternative method that requires a sensible amount of prompts for each BN and also does not assume the presence of significant information about its structure. We disregard \cite{long2023can,vashishtha2023causal} because the prompts proposed in these papers have to be used for each possible pair and triplet of BN nodes correspondingly, which seems impractical. Furthermore, we disregard \cite{long2023causal}, because it assumes the presence of preliminary information about BN structure obtained via some causal discovery algorithm. 

Thus, for further comparative experiments, we chose the prompting strategy proposed in~\cite{ban2023query}. The strategy consists of three prompts. The first prompt (\textit{Understand}) specifies the domain area, shows the node names and the values corresponding to them, and requests LLM to understand the meaning of all nodes. The second prompt (\textit{Causal discovery}) requests to analyze the cause-and-effect relationships between the nodes. The last prompt (\textit{Revision}) shows LLM the structure obtained from the response to the second prompt and asks it to revise itself. All prompts and LLM answers to them are supposed to be kept in conversation history, so LLM can reuse the information elicited during previous steps. We will refer to this method as \textit{Harness} (by the name of the paper~\cite{ban2023query}). Refer to Appendix~\ref{sec:harnes} for the exact prompts used in this method.

\subsection{Language Models}

In our experiments, we use proprietary GPT-3.5 (gpt-3.5-turbo-1106)~\cite{ouyang2022training} and GPT-4 (gpt-4-0613)~\cite{achiam2023gpt} and open-source \href{https://huggingface.co/meta-llama/Llama2-70b-chat-hf}{Llama2 70b}~\cite{touvron2023llama}. 


\subsection{Data contamination test}
\label{sec:contam_methodology}

To the best of our knowledge, there is no specific test designed to evaluate the degree of contamination of BN structure for certain LLMs (i.e., how much information about a particular BN and its structure has been seen by the LLM during its training). Thus, we design a simple approach 
inspired by the prompt technique proposed in~\cite{golchin2023time}. In this work, the authors employ a ``guided instruction'': a prompt that points to the particular piece of data to be tested for contamination and asks LLM to perform some tasks that assume that this particular piece of data has been seen by LLM during its training.

Our approach includes two prompts consecutively sent to LLM. The first prompt points to the definite paper where the BN was originally presented and to the URL address of the BN where it is publicly available (for ``popular'' BNs) and asks to generate the nodes of the BN. The second prompt requests to retrieve the edges that connect the elicited nodes. The full prompts are available in Appendix~\ref{sec:app_contam_prompts}.

\subsection{Evaluation metrics}

We evaluate the quality of the BN structure elicited from LLMs using the following metrics. First, we use Structural Hamming Distance (SHD) - a common metric used to evaluate graph discovery algorithms~\cite{tsamardinos2006max}. Low values of SHD correspond to the better quality of the learned graph. It yields a real number that results from the sum of addition, removal, or reverse edge direction operations necessary to obtain the target graph from the learned graph. The badly oriented edges (where cause and effect are mixed) are counted as two mistakes. We report SHD normalized by BN edge count to make the comparison of the algorithms more meaningful because the SHD increases with the increase in BN's size. We also analyze FP (false positives) indicating the number of edges to be deleted from a learned graph and FN (false negatives) indicating the number of edges to be added to a learned graph to get the structure of a real graph. Second, we retrieve the adjacency matrix (i.e. the matrix of 0 and 1, where 1 corresponds to the edge from one node to another) from elicited and target BNs' structures and calculate the F-score-macro between them.

\section{Results}
\label{sec:results}

\subsection{Ambiguous node names}

Whereas the BNs easily available for reuse from such websites as bnlearn are frequently used for data-driven BNSL papers~\cite{wang2021learning,behjati2020improved}, some of these BNs seem to be impractical for the tests with LLM-driven BNSL. One important problem with such BNs is that their node names are represented as ambiguous acronyms or meaningless abbreviations. Our experiments showed that the node names of \texttt{barley} and \texttt{pathfinder} BNs are impossible to decipher using LLM (with \textit{Understand} prompt of \textit{Harness} approach). Thus, during deciphering multiple nodes could be understood equally by LLM. For example, in \texttt{barley} five different nodes (\textit{nmin}, \textit{jordn}, \textit{ngodnt}, \textit{ngodnn}, and \textit{ngodn}) were deciphered as \textit{nitrogen content in the soil used for beer ingredient production}. In the case of \texttt{pathfidner} the nodes were impossible to decipher because they are represented as abbreviature of the form \textit{F1, F2, ...}. Thus, we exclude these two BNs from further experiments.

\begin{table}[]
\small
\begin{adjustbox}{width=\linewidth}
\begin{tabular}{|l|l|ll|ll|ll|}
\hline
\multirow{2}{*}{\textbf{BN}} & \multirow{2}{*}{\textbf{True\#}} & \multicolumn{2}{l|}{\textbf{Llama2}} & \multicolumn{2}{l|}{\textbf{GPT-3.5}} & \multicolumn{2}{l|}{\textbf{GPT-4}} \\ \cline{3-8} 
 &  & \multicolumn{1}{l|}{\#} & Rec & \multicolumn{1}{l|}{\#} & Rec & \multicolumn{1}{l|}{\#} & Rec \\ \hline
coma & 5 & \multicolumn{1}{l|}{35} & 0.2 & \multicolumn{1}{l|}{7} & 0 & \multicolumn{1}{l|}{0} & 0 \\ \hline
covid & 20 & \multicolumn{1}{l|}{27} & 0.05 & \multicolumn{1}{l|}{9} & 0.1 & \multicolumn{1}{l|}{0} & 0 \\ \hline
insurance & 27 & \multicolumn{1}{l|}{9} & 0.04 & \multicolumn{1}{l|}{7} & 0 & \multicolumn{1}{l|}{17} & 0.52* \\ \hline
alarm & 37 & \multicolumn{1}{l|}{15} & 0.03 & \multicolumn{1}{l|}{10} & 0.24* & \multicolumn{1}{l|}{37} & 0.86* \\ \hline
hailfinder & 56 & \multicolumn{1}{l|}{14} & 0.11 & \multicolumn{1}{l|}{7} & 0 & \multicolumn{1}{l|}{20} & 0.04 \\ \hline
hepar2 & 70 & \multicolumn{1}{l|}{20} & 0.19 & \multicolumn{1}{l|}{12} & 0.13 & \multicolumn{1}{l|}{15} & 0.16 \\ \hline
agro & 6 & \multicolumn{1}{l|}{14} & 0 & \multicolumn{1}{l|}{10} & 0.17 & \multicolumn{1}{l|}{0} & 0 \\ \hline
sperm & 9 & \multicolumn{1}{l|}{14} & 0 & \multicolumn{1}{l|}{9} & 0 & \multicolumn{1}{l|}{0} & 0 \\ \hline
screen & 16 & \multicolumn{1}{l|}{14} & 0.06 & \multicolumn{1}{l|}{7} & 0.06 & \multicolumn{1}{l|}{0} & 0 \\ \hline
sids & 17 & \multicolumn{1}{l|}{5} & 0.18 & \multicolumn{1}{l|}{8} & 0.24 & \multicolumn{1}{l|}{0} & 0 \\ \hline
apple & 29 & \multicolumn{1}{l|}{14} & 0 & \multicolumn{1}{l|}{5} & 0 & \multicolumn{1}{l|}{0} & 0 \\ \hline
urinary & 36 & \multicolumn{1}{l|}{14} & 0.03 & \multicolumn{1}{l|}{8} & 0.06 & \multicolumn{1}{l|}{0} & 0 \\ \hline
\end{tabular}
\end{adjustbox}
\caption{The results of data contamination experiments. \# corresponds to the number of generated nodes, and Rec corresponds to a recall of true nodes relative to generated nodes. The cases where the LLM used the node names that were precisely equal to the ones used in the publicly available BN are marked with a ``*'' sign.}
\label{tab:contam}
\end{table}

\subsection{Very large BNs}

The size of the BNs is another obstacle that seems interesting for analysis. As discussed in Section~\ref{sec:engaged_bns} we engage several large BNs from the bnlearn website to check how the studied LLM-driven BNSL methods perform on them. The largest BNs (\texttt{andes} with 223 nodes, and \texttt{diabetes} with 413 nodes)  exceeded the requirements of the context size available to most of the LLMs engaged in the experiments. \textit{Understand} prompt for \texttt{andes} BN did not fit into the GPT-4 and Llama 2 contexts. Whereas it fit into the GPT-3.5 context (recall, that we used gpt-3.5-turbo-1106 which had a bigger context size than gpt-4-0613), the LLM failed to decipher the meaning of all nodes. \textit{Understand} prompt for \texttt{diabetes} did not fit into the context size of any of the LLMs. \textit{Understand} prompt for \texttt{munin} BN (186 nodes) fit into the context size of all LLMs, but similarly to \texttt{andes} none of the LLMs succeeded in the deciphering of all nodes of this BN. The prompts required for running our method with the aforementioned BNs also did not fit into the context size of all LLMs. Thus, we exclude these BNs from further experiments as well.




\subsection{Data contamination}

We check the contamination of the information about each BN for each engaged LLM as discussed in Section~\ref{sec:contam_methodology} and analyze the results manually. We first check the list of nodes generated by the LLMs and map them to the node names of real BN if the sense of the generated node is equal to the real node. We analyze the obtained results by comparing the true number of the nodes in a target BN with the number of nodes generated by LLM, and also by calculating the recall of the target nodes (the split of real nodes that were generated by LLM). 

The results of the experiments (Table~\ref{tab:contam}) show that the information about any BN does not seem to be known to Llama2. However, it is very likely that \texttt{insurance} BN is known to GPT-4 and \texttt{alarm} BN is known to both GPT-3.5 and GPT-4. In the case of GPT-3.5, even though the recall of BN nodes is not very high, the way the LLM referred to the node names in its answer is precisely similar to the names used in its publicly available version on bnlearn. In the case of GPT-4, the recall for both of these BNs is very high (compared to all other cases), the LLM also uses the node names similar to the ones in publicly available BN, and in the case of \texttt{alarm} the LLM even generates the precise number of nodes included in a target BN.

Overall, it is highly probable that Llama2 and GPT-3.5 (excluding the \texttt{alarm} BN), derive their information from the name of the paper mentioned in the prompt and attempt to predict the nodes that a BN described in the mentioned paper might include. We verify this idea by prompting LLMs with the names of two non-existing papers. Llama2 and GPT-3.5 keep generating the node names similarly to what they did for the real BNs. At the same time, GPT-4 seems to be more careful in its answers. As shown in Table~\ref{tab:contam} GPT-4 has a lot of cases where no nodes were generated. In these cases, the general sense of GPT-4's reply was that the corresponding paper does not provide a detailed list of nodes BNs. In some cases (e.g., \texttt{coma}, \texttt{covid}) it generated a list of nodes that could be included in such a BN (similarly to other BNs) but it was specifically indicated in the reply that these nodes are elicited from the idea of the paper name. In such cases, we assume that no nodes have been provided. Refer to Appendix~\ref{sec:contam_answers_examples} for examples of contamination test results for some BNs.

\begin{table}[t]
\small
\centering
\begin{adjustbox}{width=\linewidth}
\begin{tabular}{|l|l|l|l|l|l|}
\hline
\textbf{LLM} & \textbf{BN} & \textbf{True \#} & \textbf{\% edg} & \textbf{F-score} & \textbf{SHD} \\ \hline
GPT-3.5 & alarm & 46 & 0.17 & 0.51 & 51 \\ \hline
GPT-4 & alarm & 46 & 0.91 & 0.63 & 60 \\ \hline
GPT-4 & insurance & 52 & 0.46 & 0.64 & 52 \\ \hline
\end{tabular}
\end{adjustbox}
\caption{The analysis of the BN structure elicited from LLM knowledge in terms of data contamination test. ``True \#'' indicates the true number of edges in the BN. ``\% edg'' indicates the ratio of the number of generated edges to the true number.}
\label{tab:contam_edges}
\end{table}

The contamination tests with other BNs yield very low recall, and the number of nodes generated by LLMs significantly differs from the true number. The cases when the recall is around 0.2 (\texttt{hepar2} and \texttt{sids}) are most likely caused by the explicit names of the papers and the awareness of the LLMs of the concepts of these papers (\texttt{hepar2} - diagnosis of liver disorders, \texttt{sids} - sudden unexpected infant death).

We also analyze the quality of the structure elicited by the LLMs in response to the second prompt of the contamination test. Such analysis makes sense only in cases where the response to the prompt about node names gives reason to suspect that the data are contaminated. Thus, we analyze the elicited structure only for three cases: \texttt{alarm} for GPT-3.5 and GPT-4 and \textit{insurance} for GPT-4 (refer to Table~\ref{tab:contam_edges}). In the case of GPT-4 for both BNs, a significant number of edges were generated, resulting in an F-score higher than 0.6, which increases the suspicions that the BN structure is compromised. In the case of GPT-3.5, the elicited structure does not seem to be sensible. 

We aim to perform further experiments only on the BNs that are unseen by all LLMs, so we exclude \texttt{alarm} and \texttt{insurance} BNs, and based on contamination test results, we assume that the rest of the BNs have not been seen by all LLMs engaged in our study. Refer to Appendix~\ref{sec:app_detailed_experiment_stat} Figure~\ref{fig:bns_selection} for the visualization of all the steps for filtering the BNs to be used for the experiments with LLM-driven BNSL.





%


\subsection{Multiple ``LLM experts''}

\begin{figure}[t]
    \centering
    \includegraphics[scale=0.4]{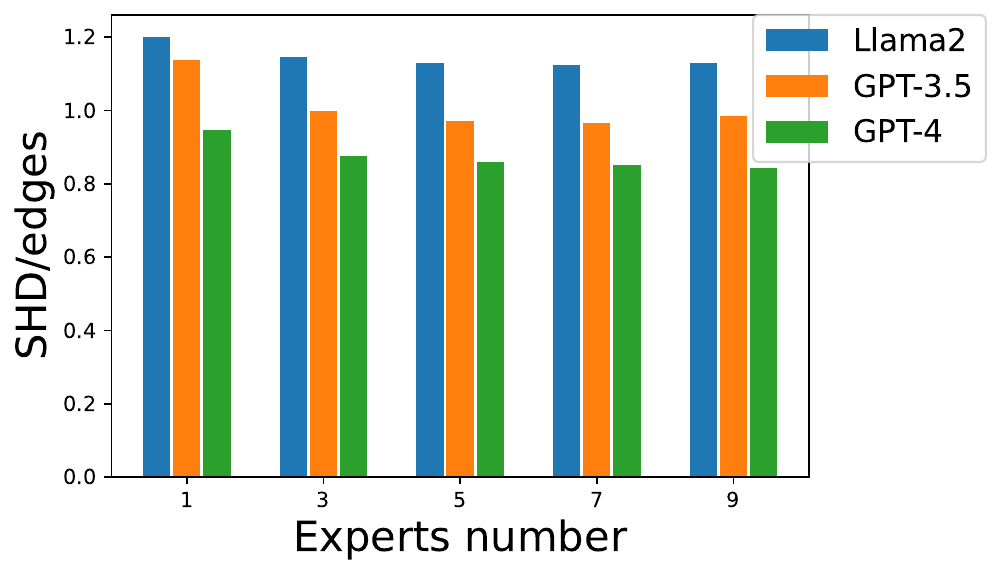}
    \caption{Dynamic of mean SHD normalized by edges count related to the number of ``LLM expert'' profiles.}
    \label{fig:experts_number}
\end{figure}

The setup of our method of LLM usage for BNSL has a hyperparameter that needs to be studied and fixed for further experiments - the number of experts. 


To study the effect of the number of expert profiles generated by the facilitator, we launch our method on all BNs that have not been previously excluded from the experiments with odd numbers of experts from 1 to 9 and calculate the mean SHD from all BNs with the corresponding expert profiles. Figure~\ref{fig:experts_number} shows that there is a positive effect of multiple expert profiles (recall that lower SHD corresponds to the better quality of the learned graph), i.e., three or more experts perform better than a single one. However, there is no best number of experts. For uniformity of further experiments, we fixed the number of experts to 7, because this seems optimal to engage a significant number of different ``opinions'' from different ``LLM experts'' and, at the same time, it gives multiple results for every BN (i.e., 36 combinations of experts), so we may check the robustness of the proposed method w.r.t. different ``LLM expert'' profiles.

\begin{figure}[t]
    \centering
    \includegraphics[scale=0.5]{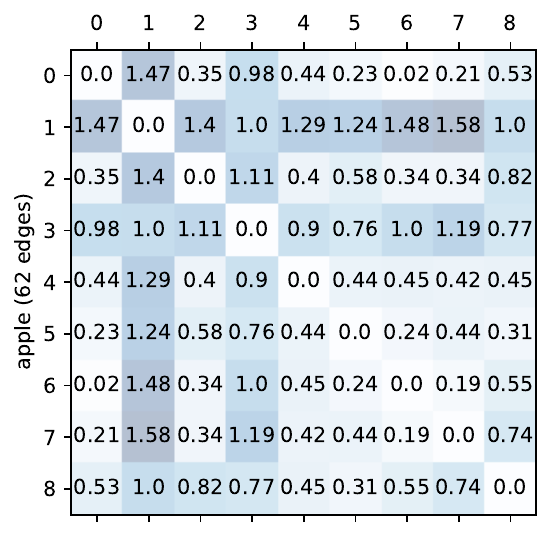}
    \caption{SHD normalized by edges count between structures generated by different GPT-4 ``LLM-experts'' for \texttt{apple} BNs.}
    \label{fig:experts_structure_diff}
\end{figure}

\begin{table}[]
\small
\centering
\begin{tabular}{|c|c|c|c|c|}
\hline
\textbf{BN} & \textbf{\# edges} & \textbf{Llama2} & \textbf{GPT-3.5} & \textbf{GPT-4} \\ \hline
coma & 5 & 0.67 & 0.61 & 0.73 \\ \hline
agro & 10 & 0.24 & 0.44 & 0.5 \\ \hline
sperm & 11 & 0.57 & 0.8 & 0.09 \\ \hline
screen & 21 & 0.4 & 0.78 & 0.4 \\ \hline
covid & 26 & 0.81 & 0.47 & 0.22 \\ \hline
sids & 27 & 0.47 & 0.42 & 0.42 \\ \hline
apple & 62 & 0.29 & 0.55 & 0.63 \\ \hline
hailfinder & 66 & 0.1 & 0.54 & 0.29 \\ \hline
urinary & 107 & 0.42 & 0.37 & 0.2 \\ \hline
hepar2 & 123 & 0.51 & 1.01 & 0.18 \\ \hline
\textit{mean} & \textit{} & \textit{0.45} & \textit{0.6} & \textit{0.37} \\ \hline
\end{tabular}
\caption{Mean SHD normalized by edges number between the BN structures proposed by 9 ``LLM-experts''.}
\label{tab:shd_between_experts}
\end{table}


We also check how the structures proposed by different instances of LLMs differ from each other. Figure~\ref{fig:experts_structure_diff} shows that BN structures generated by 9 ``LLM-experts'' for \texttt{apple} BN have significant differences. Table~\ref{tab:shd_between_experts} reporting the mean SHD normalized by edge count of each confusion matrix of this type obtained for all LLMs and BNs shows that the structures proposed by different LLM instances differ from each other and the difference varies: GPT-3.5 generates the most diverse BN structures, whereas multiple GPT-4 ``LLM experts'' tend to generate more uniform structures. See the detailed overview of inter-experts SHD for all LLMs and BNs in Appendix~\ref{sec:app_detailed_experiment_stat} Figure~\ref{fig:experts_structure_diff_all_part1} and \ref{fig:experts_structure_diff_all_part2}.

\subsection{Comparison of the methods}


\begin{table}[t]
\small
\centering
\begin{adjustbox}{width=\linewidth}
\begin{tabular}{|l|l|ll|ll|ll|}
\hline
\multicolumn{1}{|c|}{\multirow{2}{*}{\textbf{BN}}} & \multicolumn{1}{c|}{\multirow{2}{*}{\textbf{\#edg}}} & \multicolumn{2}{c|}{\textbf{Llama2}} & \multicolumn{2}{c|}{\textbf{GPT-3.5}} & \multicolumn{2}{c|}{\textbf{GPT-4}} \\ \cline{3-8} 
\multicolumn{1}{|c|}{} & \multicolumn{1}{c|}{} & \multicolumn{1}{l|}{harn} & our & \multicolumn{1}{l|}{harn} & our & \multicolumn{1}{l|}{harn} & our \\ \hline
coma & 5 & \multicolumn{1}{l|}{1.6} & 1 & \multicolumn{1}{l|}{0.4} & 0.6 & \multicolumn{1}{l|}{0.4} & 0 \\ \hline
agro & 10 & \multicolumn{1}{l|}{0.8} & 1.1 & \multicolumn{1}{l|}{1.1} & 1.2 & \multicolumn{1}{l|}{1.1} & 1 \\ \hline
sperm & 11 & \multicolumn{1}{l|}{1.27} & 0.91 & \multicolumn{1}{l|}{1.18} & 1.09 & \multicolumn{1}{l|}{1} & 0.73 \\ \hline
screen & 21 & \multicolumn{1}{l|}{1.24} & 1.33 & \multicolumn{1}{l|}{1.48} & 1.1 & \multicolumn{1}{l|}{1.29} & 1.05 \\ \hline
covid & 26 & \multicolumn{1}{l|}{0.65} & 1.15 & \multicolumn{1}{l|}{0.69} & 0.85 & \multicolumn{1}{l|}{0.58} & 0.58 \\ \hline
sids & 27 & \multicolumn{1}{l|}{1} & 1.04 & \multicolumn{1}{l|}{1.44} & 0.96 & \multicolumn{1}{l|}{1.15} & 1.22 \\ \hline
apple & 62 & \multicolumn{1}{l|}{1.02} & 1.37 & \multicolumn{1}{l|}{1.15} & 0.85 & \multicolumn{1}{l|}{0.85} & 0.94 \\ \hline
hailfinder & 66 & \multicolumn{1}{l|}{1.05} & 1.33 & \multicolumn{1}{l|}{1.24} & 1.05 & \multicolumn{1}{l|}{1.03} & 1.03 \\ \hline
urinary & 107 & \multicolumn{1}{l|}{1.01} & 1.01 & \multicolumn{1}{l|}{1.97} & 0.94 & \multicolumn{1}{l|}{0.99} & 1.01 \\ \hline
hepar2 & 123 & \multicolumn{1}{l|}{1.07} & 1.13 & \multicolumn{1}{l|}{1.17} & 1.06 & \multicolumn{1}{l|}{1.04} & 1 \\ \hline
mean &  & \multicolumn{1}{l|}{1.07} & 1.14 & \multicolumn{1}{l|}{1.18} & 0.97 & \multicolumn{1}{l|}{0.94} & 0.86 \\ \hline
\end{tabular}
\end{adjustbox}
\caption{SHD normalized by edges count of \textit{Harness} and our method.}
\label{tab:method_comparison_all_bns}
\end{table}

Table~\ref{tab:method_comparison_all_bns} shows the SHD normalized by the number of edges resulting from the two methods. We apply the Wilcoxon signed-rank test, which shows, that there is no significant difference in the performance of the methods with Llama2 and GPT-4 (p-values of 0.43 and 0.17, respectively), whereas our method performs significantly better than \textit{Harness} with GPT-3.5 (p-value 0.04).

\begin{figure}[t]
    \centering
    \includegraphics[scale=0.6]{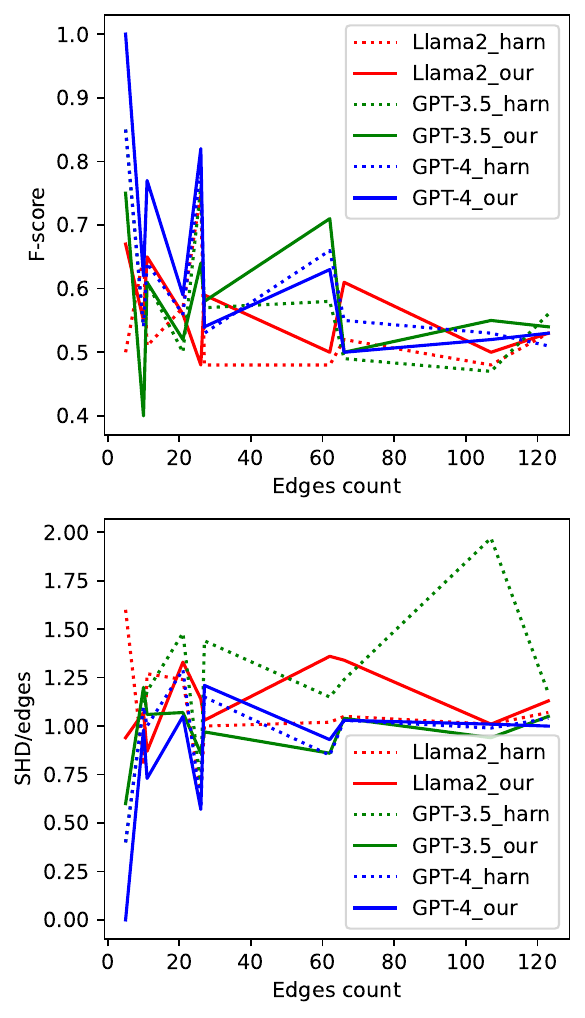}
    \caption{F-score and SHD normalized by edges count related to the number of edges in a BN.}
    \label{fig:fscore_shd_by_edges}
\end{figure}

We show more detailed statistics of both methods with GPT-4 in Table~\ref{tab:gpt4_expanded_metrics} (similar statistics for GPT-3.5 and Llama2 can be found in Appendix~\ref{sec:app_detailed_experiment_stat} Table~\ref{tab:gpt35_and_llama2_expanded_metrics}), where we see that the false negatives (i.e., the necessary edges that have not been included in the final graph) make the greatest contribution to SHD for both methods. It seems, that even though the sensible causal connections could be elicited by the best-performing GPT-4, it still cannot cover them all, especially for big BNs. Moreover, the F-score has a drastic decrease when the size of BN increases - experiments with \texttt{sids} (17 nodes, 27 edges) and bigger BNs result in a pretty low F-score, which suggests that the overall quality of the results for BNs of such size drops significantly. This could also be seen in Figure~\ref{fig:fscore_shd_by_edges}, where we show the dynamics of SHD and F-score by the number of edges, and Appendix~\ref{sec:app_detailed_experiment_stat} Figure~\ref{fig:fscore_shd_by_nodes} by nodes. All combinations of methods and LLMs do not seem capable of generating the BN structure with SHD significantly less than the number of edges. The F-score begins to border on random guessing for all BNs with more than 60 edges for all combinations of LLMs and methods as well. 


\begin{table}[t]
\small
\centering
\begin{tabular}{|c|l|l|l|l|l|}
\hline
\multicolumn{1}{|c|}{\textbf{BN}} & \textbf{Method} & \textbf{FP} & \textbf{FN} & \textbf{F-score} & \textbf{SHD/edg} \\ \hline
\multirow{2}{*}{coma} & har & 0 & 2 & 0.85 & 0.4 \\ \cline{2-6} 
 & our & 0 & \textbf{0} & \textbf{1} & \textbf{0} \\ \hline
\multirow{2}{*}{agro} & har & \textbf{3} & 8 & 0.54 & 1.1 \\ \cline{2-6} 
 & our & 4 & \textbf{6} & \textbf{0.62} & \textbf{0.98} \\ \hline
\multirow{2}{*}{sperm} & har & 3 & 8 & 0.64 & 1 \\ \cline{2-6} 
 & our & 3 & \textbf{5} & \textbf{0.77} & \textbf{0.73} \\ \hline
\multirow{2}{*}{screen} & har & 9 & 18 & 0.56 & 1.29 \\ \cline{2-6} 
 & our & \textbf{4} & 18 & \textbf{0.59} & \textbf{1.05} \\ \hline
\multirow{2}{*}{covid} & har & \textbf{0} & 15 & 0.79 & 0.58 \\ \cline{2-6} 
 & our & 2 & \textbf{12} & \textbf{0.82} & 0.57 \\ \hline
\multirow{2}{*}{sids} & har & \textbf{6} & 25 & 0.53 & \textbf{1.15} \\ \cline{2-6} 
 & our & 8 & \textbf{24} & \textbf{0.54} & 1.21 \\ \hline
\multirow{2}{*}{apple} & har & \textbf{5} & \textbf{48} & \textbf{0.66} & \textbf{0.85} \\ \cline{2-6} 
 & our & 8 & 49 & 0.63 & 0.93 \\ \hline
\multirow{2}{*}{hailfinder} & har & 6 & \textbf{62} & \textbf{0.55} & 1.03 \\ \cline{2-6} 
 & our & \textbf{2} & 66 & 0.5 & 1.03 \\ \hline
\multirow{2}{*}{urinary} & har & \textbf{5} & \textbf{101} & \textbf{0.53} & \textbf{0.99} \\ \cline{2-6} 
 & our & 6 & 102 & 0.52 & 1.01 \\ \hline
\multirow{2}{*}{hepar2} & har & 7 & 121 & 0.51 & 1.04 \\ \cline{2-6} 
 & our & \textbf{5} & \textbf{118} & \textbf{0.53} & \textbf{1} \\ \hline
\end{tabular}
\caption{Detailed information about the performance of \textit{Harness} and our method with GPT-4. FP and FN correspond to false positive and false negative edges in the learned BN structure.}
\label{tab:gpt4_expanded_metrics}
\end{table}

\section{Error analysis}
\label{sec:error_analysis}

In this section, we discuss several errors found typical for different LLMs applied for both studied methods. The examples of all errors can be found in the Appendix~\ref{sec:appendix_error_analysis}.

Some LLMs, in particular, GPT-3.5 are prone to generating cycles (i.e., two edges of the graph directed in different directions between \textit{node A} and \textit{node B}). It turns out to be most critical for our method, whereas \textit{Harness} method generates almost no cycles with all LLMs. See the detailed statistics of cycles in  Appendix~\ref{sec:appendix_error_analysis} Tables~\ref{tab:cycles_average} and ~\ref{tab:duplicates_expanded}.


The ability to stick to the node names proposed for creating the BN structure is another problem that yields errors in our method. Recall that we use the maximally unambiguous names of the nodes, explained with \textit{Understand} prompt of \textit{Harness} method. On the one hand, it gives the LLM a clear input, but on the other hand, when the LLM generates a final JSON, it may fail to accurately reproduce all the names of the nodes. Sometimes such errors are related to such nonsensible differences as capitalization changes. In such cases, we try to find the most similar name of the node using Levenshtein distance. If no matches are found, we have to manually analyze the output and decide what to do with the nodes, whose names are significantly different from any nodes provided to create a BN structure. It turns out that LLMs may ``invent'' the nodes. Sometimes, it may take the node name either from the system message or its reasoning (answer to the first prompt), where new terminology appears. For example, Llama2 instantiated as \textit{Epidemiologist with 15 years of experience in studying the distribution and determinants of health-related events, diseases, or health-related characteristics among populations, including the epidemiology of \textbf{urinary tract infections} in children} generated a BN structure of \texttt{urinary} BN with most edges connected to a node \textit{Urinary tract infection}, which was not proposed as a possible node name in the prompts.


Sometimes LLM may refuse to generate any structure because of ethical or some other reasons (e.g. \textit{``As a medical ethicist, I am not qualified to determine the direct relationships between these medical factors.''}). In such cases, we interpret the resulting JSON as an empty list.
 



Finally, in some cases, LLM seems to misunderstand the idea of causal relations typical for BN structure, and instead of complex analysis, it may start generating a cause-and-effect chain.



\section{Discussion}
\label{sec:discussion}



In this section, we discuss the main insights of our study.

First, the proper results of LLM usage for BNSL require proper input. Whereas in the case of data-driven algorithms, the meaning of nodes is not important, for LLM-driven BNSL, the maximally disambiguated sense of each node is crucial because it is to be used for further reasoning about causal relationships between the nodes. There is no way to control the clarity of BN nodes when we test the algorithm on the existing BNs. For using any of the discussed algorithms for the real task, the clarity of the nodes turns out to be crucial. Moreover, in the case of using our approach, we experimented with LLM-generated expert profiles, which is another point to be controlled. The most optimal way could be to have a domain specialist capable of unambiguous definitions of all input values for launching the BN structure elicitation: node names, knowledge area, and profiles of ``LLM experts''.

Second, it is clear that the performance of both of the discussed methods significantly decreases when the number of nodes and edges in BNs increases. In some cases, the problems were caused by the context size of the LLM selected for the experiments. This could be solved with LLMs with bigger context (e.g. gpt-4-32k), but the experiments show that the increase of LLM context size is unlikely to solve the main problem, which seems to be the inability of all LLMs engaged in the experiments to reason at length about the relationships between a large number of factors without losing detail. The used LLMs seem to be capable of identifying only key relationships between factors.

\section{Conclusion}

In this work, we proposed a novel method of LLM usage for BN structure learning, which initializes several LLMs with different experiences, independently queries them about the structure of the BN, and obtains the final structure of the BN by majority voting. We study the performance of this and another method on widely and not widely known BNs of different sizes. Moreover, we proposed a simple approach to check the contamination of BN in LLM, which shows that some widely known BNs are unsuitable for testing the LLM usage for BN structure elicitation. Our experiments showed that certain widely known BNs are also unsuitable for usage with such structure learning methods because their nodes are impossible to understand or their size is too big for LLM context size. The experiments on the other BNs show that our method performs better than the existing method with one of the three studied LLMs; however, both of the methods suffer a significant drop in performance when the size of BNs increases.




\bibliography{main}
\bibliographystyle{acl_natbib}

\appendix

\section{Our method overview}
\label{sec:app_our_method}

\subsection{Facilitator prompts}
\label{sec:app_our_method_facilitator}

\textbf{System message}

We are going to collect a Bayesian Network using a special communication protocol. The protocol is based on the paper "BARD: A Structured Technique for Group Elicitation of Bayesian Networks to Support Analytic Reasoning". It assumes that several specialists possess the necessary skills in the Bayesian Networks problem domain, and respond to our questions independently. Then we match their responses and help them to discuss the answers in an anonymous mode if any disagreements are found until a collective agreement is achieved.

\textbf{First prompt requesting to think about possible profiles of the experts}

 We are going to collect a Bayesian network that requires some knowledge about \textit{knowledge area}      Here is the general idea of the Bayesian Network: \textit{main task of the BN}. We will use another Large Language Model as experts. We will need 9 profiles of the experts that will be used to initialize the system message of the Language Model. The profiles must be as diverse as possible but at the same time, they must jointly possess all necessary knowledge to fulfill the task of knowledge elicitation for Bayesian Network collection. Think step-by-step what are the main qualities such experts should possess.

 \textbf{Second prompt requesting to generate a valid JSON with the profiles of the experts}

 Now please propose to me 9 profiles of the experts that will be used to initialize the system message of the Language Model. Turn your answer into JSON of the following form. Obligatory use such json from and do not include any side comments  ```json      \{"expert\_1": "textual description of expert" (simply copy paste the details you used in the previous reply), "expert\_2": "textual description of expert"(simply copy paste the details you used in the previous reply), "expert\_3": "textual description of expert" (simply copy paste the details you used in the previous reply)          ...       \}  ``` 

 \subsection{LLM expert prompts}
 \label{sec:app_our_method_experts}

 \textbf{System message}

 You will generate a predictive model using a specialized communication protocol. Assume the presence of multiple specialists possessing the required skills in the designated problem domain. Each specialist responds independently to our questions. Provide input as an expert with the following profile: \textit{profile of the expert}.

 \textbf{First prompt demonstrating the list of explicit names of the BN nodes and requesting to reason about possible causal relations between them.}

 Consider the factors associated with the predictive model, represented by the list of nodes:\textit{list of explicit names of BN nodes}. Now, analyze the relationships between these factors. 
 
 There are three possible types of relations: 
 \begin{itemize}
     \item Factor A directly affects Factor B
     \item Factor B directly affects Factor A
     \item No direct effect between the two factors
 \end{itemize}
 
 Please systematically evaluate the interconnections between the specified factors, focusing only on significant relations

 \textbf{Second prompt requesting to summarize the generated causal relationships into a valid JSON}

 Summarize your thoughts in valid JSON format based on the relationships between the specified factors: \textit{list of explicit names of BN nodes}. Use the following format to indicate connections between factors A and B: [(factor A, factor B)] (indicating that A directly affects B). Obligatorily keep the original names of the specified factors, do not change any letter from them. Provide only the valid JSON representation without additional discussion, following this structure:
 
 [ [factor A, factor B], (meaning the factor A directly affects factor B)
 
 [factor C, factor E], (meaning the factor C directly affects factor E)
 
 [factor D, factor H], (meaning the factor D directly affects factor H)
 
 .....
 [factor .., factor ...]]

 \textbf{Decycling prompt}

 You have indicated that the changes of factor \textit{node A} directly causally affect the changes of factor \textit{node B} and the changes of factor \textit{node B} directly causally affect the changes of factor \textit{node A}  in the same time. Please note that this is not possible in terms of the model that we develop. Consider possible ways to address this issue, determining which factor has a stronger effect on the other. Systematically think through the resolution and at the end summarize your thoughts in valid JSON of one of the following forms:
 
 If you believe that the changes of factor  \textit{node A}  directly causally affect the changes of factor  \textit{node B}  more than the changes of factor  \textit{node B} directly causally affect the changes of factor  \textit{node A}  output the JSON of form
 
 ```json [ \textit{node A} ,  \textit{node B} ] 
 ```                  
 
 OR
 
 If you believe that the changes of factor \textit{node B} directly causally affect the changes of factor \textit{node A} more than the changes of factor \textit{node A} directly causally affect the changes of factor \textit{node B} output the JSON of form
 
 ```json [ \textit{node B} ,  \textit{node A} ] 
 ```

\section{Harness method overview}
 \label{sec:harnes}

\textbf{Prompt Understand}

You are an expert on Domain. You are investigating the cause and effect relationships between the following variables in your filed. Variable abbreviations and values are presented as follows.

Please understand the real meaning of each variable according to their possible values, and explain them in order.

variable sym1, values val1
· · ·
variable symn, values valn

\textbf{Prompt Causal Discovery:}

Based on the meaning of variables you provide, analyze the cause-and-effect relationships between them. Please give the results as a directed graph network in the form of <edge>variable1→variable2<\\edge>. Make sure that each edge represent a direct causality between the two variables.

\textbf{Prompt Revsion:}

Based on your explanation, check whether the following causal
statements are correct, and give the reasons.

xi1 → xj1
· · ·
xim → xjm

\section{Data contamination}
\label{sec:app_contam}

\subsection{Prompts}
\label{sec:app_contam_prompts}

\textbf{Node names request}

Generate a list of nodes in the Bayesian Network discussed in the paper \textit{paper name} and also available on \textit{URL}. Provide details on each node and its role within the network structure.

\textbf{BN structure request}

Now retrieve the edges connecting the previously mentioned nodes in the Bayesian Network described in paper \textit{paper name} and also available on \textit{URL}. Express the network structure using the 'A->B' notation, indicating the presence of an edge from node A to node B in the Bayesian Network.

\subsection{Example of answers}
\label{sec:contam_answers_examples}

\subsubsection{Answers about alarm BN}

\textbf{GPT-4 and alarm BN:} \textit{the data is very likely to be contaminated, because the answer covers almost all the nodes from bnlearn website and the names of the nodes are almost always similar to the ones on bnlearn.}

\textit{User ptompt}

 Generate a list of nodes in the Bayesian Network discussed in paper "I. A. Beinlich, H. J. Suermondt, R. M. Chavez, and G. F. Cooper. The ALARM Monitoring System: A Case Study with Two Probabilistic Inference Techniques for Belief Networks. In Proceedings of the 2nd European Conference on Artificial Intelligence in Medicine, pages 247-256. Springer-Verlag, 1989." and also available on bnlearn.com/bnrepository/discrete-medium.html. Provide details on each node and its role within the network structure.

\textit{System reply}

The ALARM (A Logical Alarm Reduction Mechanism) network is a Bayesian network that represents a model of patient monitoring in intensive care units. It consists of 37 nodes and 46 edges. Here is a list of the nodes and their roles within the network:

1. History: Represents the patient's history of lung diseases.

2. CVP: Central Venous Pressure, a measure of blood pressure in the heart.

3. PCWP: Pulmonary Capillary Wedge Pressure, a measure of pressure in the heart.

4. Hypovolemia: A condition of decreased blood volume.

5. LVEDVolume: Left Ventricular End-Diastolic Volume, a measure of the amount of blood in the heart.

6. StrokeVolume: The amount of blood pumped by the left ventricle of the heart in one contraction.

7. CO: Cardiac Output, the amount of blood the heart pumps through the circulatory system in a minute.

8. BP: Blood Pressure.

9. TPR: Total Peripheral Resistance, the resistance blood must overcome to be able to move within the body.

10. Anaphylaxis: A severe, potentially life-threatening allergic reaction.

11. InsuffAnesth: Insufficient Anesthesia.

12. TPR: Total Peripheral Resistance.

13. Anaphylaxis: Severe allergic reaction.

14. MinVol: Minimum Volume.

15. FIO2: Fraction of Inspired Oxygen, the concentration of oxygen in the inhaled air.

16. PVSat: Peripheral Venous Saturation, a measure of the oxygen saturation in the peripheral veins.

17. SaO2: Arterial Oxygen Saturation, a measure of the amount of oxygen-carrying hemoglobin in the blood.

18. PulmShunt: Pulmonary Shunt, a condition that allows blood to bypass the lungs.

19. Intubation: The process of inserting a tube through the mouth and then into the airway.

20. Press: Pressure.

21. Disconnect: Disconnection of the patient from the ventilator.

22. MinVolSet: Minimum Volume Set.

23. VentMach: Ventilation Machine.

24. KinkedTube: Kinked Tube.

25. VentTube: Ventilation Tube.

26. VentLung: Ventilation Lung.

27. VentAlv: Ventilation Alveoli.

28. ArtCO2: Arterial CO2 Pressure.

29. CO2: Carbon Dioxide.

30. HR: Heart Rate.

31. Catechol: Catecholamines, hormones made by the adrenal glands.

32. HRBP: Heart Rate Blood Pressure.

33. ErrLowOutput: Error Low Output.

34. HREKG: Heart Rate EKG.

35. HRSat: Heart Rate Oxygen Saturation.

36. InsuffAnesth: Insufficient Anesthesia.

37. ExpCO2: Expired CO2.

Each node represents a medical condition, a medical measurement, or a medical procedure. The edges between the nodes represent causal relationships between these conditions, measurements, and procedures. For example, the node "Hypovolemia" (decreased blood volume) has edges to nodes "CVP" (Central Venous Pressure) and "PCWP" (Pulmonary Capillary Wedge Pressure), indicating that hypovolemia can cause changes in these measurements.

\textit{User ptompt}

 Now retrieve the edges connecting the previously mentioned nodes in the Bayesian Network described in paper "I. A. Beinlich, H. J. Suermondt, R. M. Chavez, and G. F. Cooper. The ALARM Monitoring System: A Case Study with Two Probabilistic Inference Techniques for Belief Networks. In Proceedings of the 2nd European Conference on Artificial Intelligence in Medicine, pages 247-256. Springer-Verlag, 1989." and also available on bnlearn.com/bnrepository/discrete-medium.html. Express the network structure using the 'A->B' notation, indicating the presence of an edge from node A to node B in the Bayesian Network.

\textit{System reply}

The ALARM network is a complex Bayesian network with 46 edges. Here are the edges connecting the nodes in the network, expressed in the 'A->B' notation:

1. History->LVEDVolume

2. History->CVP

3. History->PCWP

4. CVP->LVEDVolume

5. PCWP->LVEDVolume

6. Hypovolemia->CVP

7. Hypovolemia->PCWP

8. LVEDVolume->StrokeVolume

9. StrokeVolume->CO

10. Hypovolemia->CO

11. CO->BP

12. TPR->BP

13. Anaphylaxis->TPR

14. Anaphylaxis->InsuffAnesth

15. InsuffAnesth->TPR

16. TPR->BP

17. Anaphylaxis->MinVol

18. MinVol->FIO2

19. FIO2->PVSat

20. PVSat->SaO2

21. SaO2->PulmShunt

22. PulmShunt->Intubation

23. Intubation->Press

24. Press->Disconnect

25. Disconnect->MinVolSet

26. MinVolSet->VentMach

27. VentMach->KinkedTube

28. KinkedTube->VentTube

29. VentTube->VentLung

30. VentLung->VentAlv

31. VentAlv->ArtCO2

32. ArtCO2->CO2

33. CO2->HR

34. HR->Catechol

35. Catechol->HRBP

36. HRBP->ErrLowOutput

37. ErrLowOutput->HREKG

38. HREKG->HRSat

39. HRSat->InsuffAnesth

40. InsuffAnesth->ExpCO2

41. ExpCO2->KinkedTube

42. KinkedTube->MinVolSet

43. MinVolSet->VentMach

44. VentMach->VentTube

45. VentTube->VentLung

46. VentLung->VentAlv

Each edge represents a causal relationship between two nodes in the network. For example, the edge 'Hypovolemia->CVP' indicates that hypovolemia (decreased blood volume) can cause changes in CVP (Central Venous Pressure).

\textbf{GPT-3.5 and alarm BN:} \textit{the data is also likely to be contaminated because the names of the nodes are similar to the ones on bnlearn website}

\textit{User prompt}

Generate a list of nodes in the Bayesian Network discussed in paper "I. A. Beinlich, H. J. Suermondt, R. M. Chavez, and G. F. Cooper. The ALARM Monitoring System: A Case Study with Two Probabilistic Inference Techniques for Belief Networks. In Proceedings of the 2nd European Conference on Artificial Intelligence in Medicine, pages 247-256. Springer-Verlag, 1989." and also available on bnlearn.com/bnrepository/discrete-medium.html. Provide details on each node and its role within the network structure.

\textit{System reply}

 The Bayesian Network discussed in the paper "The ALARM Monitoring System: A Case Study with Two Probabilistic Inference Techniques for Belief Networks" consists of the following nodes:

1. HISTORY
   - Role: Represents the patient's history of similar conditions.
   - Description: This node captures the patient's medical history and is used to assess the likelihood of the current condition based on past occurrences.

2. CVP
   - Role: Represents the central venous pressure.
   - Description: This node captures the patient's central venous pressure, which is an important indicator of the patient's cardiovascular status.

3. PCWP
   - Role: Represents the pulmonary capillary wedge pressure.
   - Description: This node captures the patient's pulmonary capillary wedge pressure, which is an important indicator of the patient's cardiac function.

4. HYPOTENSION
   - Role: Represents the presence of hypotension.
   - Description: This node captures the presence of hypotension in the patient, which is a critical condition that requires immediate attention.

5. LVEDVOLUME
   - Role: Represents the left ventricular end-diastolic volume.
   - Description: This node captures the patient's left ventricular end-diastolic volume, which is an important indicator of the patient's cardiac function.

6. LVFAILURE
   - Role: Represents the presence of left ventricular failure.
   - Description: This node captures the presence of left ventricular failure in the patient, which is a critical condition that requires immediate attention.

7. STROKEVOLUME
   - Role: Represents the stroke volume.
   - Description: This node captures the patient's stroke volume, which is an important indicator of the patient's cardiac function.

8. ERRLOWOUTPUT
   - Role: Represents the presence of low cardiac output.
   - Description: This node captures the presence of low cardiac output in the patient, which is a critical condition that requires immediate attention.

9. HRBP
   - Role: Represents the arterial blood pressure.
   - Description: This node captures the patient's arterial blood pressure, which is an important indicator of the patient's cardiovascular status.

10. HREKG
    - Role: Represents the heart rate from an EKG.
    - Description: This node captures the patient's heart rate as measured by an electrocardiogram, which is an important indicator of the patient's cardiac function.

These nodes collectively form the Bayesian Network used in the ALARM Monitoring System to assess the patient's cardiovascular and cardiac status and make inferences about potential critical conditions.

\textit{User prompt}

Now retrieve the edges connecting the previously mentioned nodes in the Bayesian Network described in paper "I. A. Beinlich, H. J. Suermondt, R. M. Chavez, and G. F. Cooper. The ALARM Monitoring System: A Case Study with Two Probabilistic Inference Techniques for Belief Networks. In Proceedings of the 2nd European Conference on Artificial Intelligence in Medicine, pages 247-256. Springer-Verlag, 1989." and also available on bnlearn.com/bnrepository/discrete-medium.html. Express the network structure using the 'A->B' notation, indicating the presence of an edge from node A to node B in the Bayesian Network.

\textit{System reply}

The edges connecting the nodes in the Bayesian Network described in the paper "The ALARM Monitoring System: A Case Study with Two Probabilistic Inference Techniques for Belief Networks" are as follows:

1. HISTORY -> HYPOTENSION

2. CVP -> LVFAILURE

3. PCWP -> LVFAILURE

4. HYPOTENSION -> LVFAILURE

5. LVEDVOLUME -> LVFAILURE

6. LVFAILURE -> ERRLOWOUTPUT

7. LVFAILURE -> STROKEVOLUME

8. LVFAILURE -> HREKG

9. LVFAILURE -> HREKG

10. LVFAILURE -> HRBP

These edges represent the relationships and dependencies between the nodes in the Bayesian Network, indicating the flow of probabilistic influence and information within the network structure.

\textbf{Llama2 and alarm BN:} \textit{the data is unlikely to be contaminated because the proposed nodes do not coincide with almost any of the nodes from the BN on bnlearn website} 

\textit{User prompt}

Generate a list of nodes in the Bayesian Network discussed in paper "I. A. Beinlich, H. J. Suermondt, R. M. Chavez, and G. F. Cooper. The ALARM Monitoring System: A Case Study with Two Probabilistic Inference Techniques for Belief Networks. In Proceedings of the 2nd European Conference on Artificial Intelligence in Medicine, pages 247-256. Springer-Verlag, 1989." and also available on bnlearn.com/bnrepository/discrete-medium.html. Provide details on each node and its role within the network structure.

\textit{System reply}

The Bayesian Network (BN) discussed in the paper "The ALARM Monitoring System: A Case Study with Two Probabilistic Inference Techniques for Belief Networks" by I. A. Beinlich, H. J. Suermondt, R. M. Chavez, and G. F. Cooper (1989) is a discrete Bayesian Network that models the probability distribution over 15 variables (nodes) related to the monitoring of a chemical process. The network is available on bnlearn.com/bnrepository/discrete-medium.html.

Here's a list of the nodes in the network, along with a brief description of each node and its role within the network structure:

1. Alarm: This node represents the alarm status of the chemical process. It has two possible values: "alarm" (A) and "no alarm" (not A). The probability distribution over this node is updated based on the values of the other nodes in the network.

2. Pressure: This node represents the pressure reading of a pressure sensor in the chemical process. It has three possible values: "low" (P\_L), "normal" (P\_N), and "high" (P\_H). The probability distribution over this node is updated based on the values of the other nodes in the network.

3. Flow: This node represents the flow rate of the chemical process. It has three possible values: "low" (F\_L), "normal" (F\_N), and "high" (F\_H). The probability distribution over this node is updated based on the values of the other nodes in the network.

4. Level: This node represents the level of a tank in the chemical process. It has three possible values: "low" (L), "normal" (N), and "high" (H). The probability distribution over this node is updated based on the values of the other nodes in the network.

5. Temperature: This node represents the temperature reading of a temperature sensor in the chemical process. It has three possible values: "low" (T\_L), "normal" (T\_N), and "high" (T\_H). The probability distribution over this node is updated based on the values of the other nodes in the network.

6. Conductivity: This node represents the conductivity reading of a conductivity sensor in the chemical process. It has three possible values: "low" (C\_L), "normal" (C\_N), and "high" (C\_H). The probability distribution over this node is updated based on the values of the other nodes in the network.

7. pH: This node represents the pH reading of a pH sensor in the chemical process. It has three possible values: "acidic" (pH\_A), "normal" (pH\_N), and "basic" (pH\_B). The probability distribution over this node is updated based on the values of the other nodes in the network.

8. Oxygen: This node represents the oxygen reading of an oxygen sensor in the chemical process. It has three possible values: "low" (O\_L), "normal" (O\_N), and "high" (O\_H). The probability distribution over this node is updated based on the values of the other nodes in the network.

9. Nitrogen: This node represents the nitrogen reading of a nitrogen sensor in the chemical process. It has three possible values: "low" (N2\_L), "normal" (N2\_N), and "high" (N2\_H). The probability distribution over this node is updated based on the values of the other nodes in the network.

10. Water: This node represents the water reading of a water sensor in the chemical process. It has three possible values: "low" (W\_L), "normal" (W\_N), and "high" (W\_H). The probability distribution over this node is updated based on the values of the other nodes in the network.

11. Air: This node represents the air reading of an air sensor in the chemical process. It has three possible values: "low" (A\_L), "normal" (A\_N), and "high" (A\_H). The probability distribution over this node is updated based on the values of the other nodes in the network.

12. Steam: This node represents the steam reading of a steam sensor in the chemical process. It has three possible values: "low" (S\_L), "normal" (S\_N), and "high" (S\_H). The probability distribution over this node is updated based on the values of the other nodes in the network.

13. Valve1: This node represents the status of a valve in the chemical process. It has two possible values: "open" (V1\_O) and "closed" (V1\_C). The probability distribution over this node is updated based on the values of the other nodes in the network.

14. Valve2: This node represents the status of a second valve in the chemical process. It has two possible values: "open" (V2\_O) and "closed" (V2\_C). The probability distribution over this node is updated based on the values of the other nodes in the network.

15. Pump: This node represents the status of a pump in the chemical process. It has two possible values: "on" (P\_O) and "off" (P\_C). The probability distribution over this node is updated based on the values of the other nodes in the network.

The Bayesian Network is structured such that each node is connected to other nodes in the network, representing the conditional dependencies between the variables. For example, the probability distribution over the "Alarm" node is updated based on the values of the "Pressure," "Flow," "Level," "Temperature," "Conductivity," "pH," "Oxygen," "Nitrogen," "Water," "Air," and "Steam" nodes. Similarly, the probability distribution over the "Pressure" node is updated based on the values of the "Flow," "Level," "Temperature," and "Conductivity" nodes.

The network uses a directed acyclic graph (DAG) structure to represent the conditional dependencies between the variables. The direction of the arrows in the graph indicates the direction of causality between the variables. The probability distributions over each node are updated using Bayes' rule, which combines the prior probability distribution over the node with the likelihood of the node given the values of its parent nodes in the DAG.

Overall, the Bayesian Network provides a probabilistic model of the chemical process that can be used to reason about the likelihood of different states of the process given the available sensor readings. The network can also be used to identify potential faults or anomalies in the process and to make predictions about future states of the process based on historical data.

\textit{User prompt}

Now retrieve the edges connecting the previously mentioned nodes in the Bayesian Network described in paper "I. A. Beinlich, H. J. Suermondt, R. M. Chavez, and G. F. Cooper. The ALARM Monitoring System: A Case Study with Two Probabilistic Inference Techniques for Belief Networks. In Proceedings of the 2nd European Conference on Artificial Intelligence in Medicine, pages 247-256. Springer-Verlag, 1989." and also available on bnlearn.com/bnrepository/discrete-medium.html. Express the network structure using the 'A->B' notation, indicating the presence of an edge from node A to node B in the Bayesian Network.

\textit{System reply}
  
  Sure! Here are the edges connecting the nodes in the Bayesian Network described in the paper "The ALARM Monitoring System: A Case Study with Two Probabilistic Inference Techniques for Belief Networks" by I. A. Beinlich, H. J. Suermondt, R. M. Chavez, and G. F. Cooper:

* Alarm -> Pressure

* Alarm -> Flow

* Alarm -> Level

* Alarm -> Temperature

* Alarm -> Conductivity

* Alarm -> pH

* Alarm -> Oxygen

* Alarm -> Nitrogen

* Alarm -> Water

* Alarm -> Air

* Alarm -> Steam

* Pressure -> Flow

* Pressure -> Level

* Pressure -> Temperature

* Pressure -> Conductivity

* Flow -> Level

* Flow -> Temperature

* Flow -> Conductivity

* Level -> Temperature

* Level -> Conductivity

* Temperature -> Conductivity

* Conductivity -> pH

* Conductivity -> Oxygen

* Conductivity -> Nitrogen

* Conductivity -> Water

* Conductivity -> Air

* Conductivity -> Steam

Here's the network structure using the 'A->B' notation:

* Alarm -> Pressure

* Alarm -> Flow

* Alarm -> Level

* Alarm -> Temperature

* Alarm -> Conductivity

* Alarm -> pH

* Alarm -> Oxygen

* Alarm -> Nitrogen

* Alarm -> Water

* Alarm -> Air

* Alarm -> Steam

* Pressure -> Flow

* Pressure -> Level

* Pressure -> Temperature

* Pressure -> Conductivity

* Flow -> Level

* Flow -> Temperature

* Flow -> Conductivity

* Level -> Temperature

* Level -> Conductivity

* Temperature -> Conductivity

* Conductivity -> pH

* Conductivity -> Oxygen

* Conductivity -> Nitrogen

* Conductivity -> Water

* Conductivity -> Air

* Conductivity -> Steam

Note that the network structure is a directed acyclic graph (DAG), which means that there are no cycles in the graph and the edges have a direction, indicating the direction of causality between the variables.

\subsubsection{Answers about non-existing BN}

\textbf{GPT-4 and non-existing BN:} \textit{Can not find the paper and warns that the list of nodes will be the hypothetical.}

\textit{User prompt}

 Generate a list of nodes in the Bayesian Network discussed in paper "Frozen Paths: Leveraging Bayesian Networks for Enhanced Risk Analysis in Severe Freeze Weather during Road Construction". Provide details on each node and its role within the network structure.

\textit{System reply}

 The paper "Frozen Paths: Leveraging Bayesian Networks for Enhanced Risk Analysis in Severe Freeze Weather during Road Construction" does not seem to exist, as multiple searches in academic databases yield no results. Therefore, it's impossible to provide a list of nodes in the Bayesian Network discussed in this paper. 

However, in a hypothetical Bayesian Network for risk analysis in severe freeze weather during road construction, the nodes could represent various factors affecting the construction process. Here are some potential nodes:

1. Weather Conditions: This node could represent the probability of severe freeze weather. It would be a parent node to many other nodes, as weather conditions can affect many aspects of road construction.

2. Construction Materials: This node could represent the type and quality of materials used in construction. The performance of these materials could be affected by severe freeze weather.

3. Worker Safety: This node could represent the risk of accidents or injuries to workers due to severe freeze weather.

4. Construction Equipment: This node could represent the reliability and performance of construction equipment in severe freeze weather.

5. Project Delays: This node could represent the probability of the construction project being delayed due to severe freeze weather.

6. Cost Overruns: This node could represent the probability of the construction project going over budget due to severe freeze weather.

7. Quality of Construction: This node could represent the overall quality of the construction project, which could be affected by severe freeze weather.

Each of these nodes would have various parent and child nodes, creating a complex network of dependencies and probabilities. The Bayesian Network would be used to analyze and predict the overall risk of the construction project in the face of severe freeze weather.

\textbf{GPT-3.5 and non-existing BN:} \textit{Generates the list of nodes (without any warnings)}

\textit{User prompt}

 Generate a list of nodes in the Bayesian Network discussed in paper "Frozen Paths: Leveraging Bayesian Networks for Enhanced Risk Analysis in Severe Freeze Weather during Road Construction". Provide details on each node and its role within the network structure.

\textit{System reply}

  1. Weather Conditions
   - This node represents the different weather conditions that can occur during road construction, such as severe freeze, snow, ice, etc. It plays a crucial role in determining the risk of severe freeze weather affecting the construction site.

2. Road Construction Activities
   - This node represents the various activities involved in road construction, such as laying asphalt, pouring concrete, etc. It influences the likelihood of severe freeze weather impacting the construction process.

3. Risk of Severe Freeze Weather
   - This node represents the likelihood of severe freeze weather occurring at the construction site. It is influenced by the weather conditions and other environmental factors.

4. Impact on Construction Schedule
   - This node represents the potential impact of severe freeze weather on the construction schedule. It is influenced by the risk of severe freeze weather and road construction activities.

5. Safety Risks
   - This node represents the safety risks associated with severe freeze weather during road construction. It is influenced by the impact on construction schedule and other factors.

6. Mitigation Strategies
   - This node represents the various strategies that can be implemented to mitigate the impact of severe freeze weather on road construction. It is influenced by the risk of severe freeze weather and safety risks.

7. Cost Implications
   - This node represents the potential cost implications of severe freeze weather on road construction. It is influenced by the impact on construction schedule and mitigation strategies.

These nodes collectively form a Bayesian Network that allows for enhanced risk analysis of severe freeze weather during road construction, taking into account various factors and their interdependencies.

\textbf{Llama2 and non-existing BN:} \textit{Generates the list of nodes (without any warnings)}

\textit{User prompt}

  Generate a list of nodes in the Bayesian Network discussed in paper "Frozen Paths: Leveraging Bayesian Networks for Enhanced Risk Analysis in Severe Freeze Weather during Road Construction". Provide details on each node and its role within the network structure.

\textit{System reply}

The Bayesian Network discussed in the paper "Frozen Paths: Leveraging Bayesian Networks for Enhanced Risk Analysis in Severe Freeze Weather during Road Construction" is designed to assess the risk of frozen ground during road construction in cold regions. The network consists of 14 nodes, each representing a specific variable or factor that affects the risk of frozen ground. Here is a list of the nodes, along with a brief description of each node and its role within the network structure:

1. Air Temperature (AT): Represents the temperature of the air in degrees Celsius. AT is a primary input variable that affects the risk of frozen ground.

2. Ground Temperature (GT): Represents the temperature of the ground in degrees Celsius. GT is a primary output variable that indicates the likelihood of frozen ground.

3. Soil Moisture (SM): Represents the moisture content of the soil. SM affects the rate of heat transfer between the ground and the air, and therefore affects the risk of frozen ground.

4. Wind Speed (WS): Represents the wind speed in meters per second. WS affects the rate of heat transfer between the ground and the air, and therefore affects the risk of frozen ground.

5. Relative Humidity (RH): Represents the relative humidity of the air. RH affects the rate of heat transfer between the ground and the air, and therefore affects the risk of frozen ground.

6. Cloud Cover (CC): Represents the extent of cloud cover. CC affects the amount of solar radiation that reaches the ground, which in turn affects the risk of frozen ground.

7. Time of Day (TOD): Represents the time of day, which affects the amount of solar radiation that reaches the ground, and therefore affects the risk of frozen ground.

8. Day of Year (DOY): Represents the day of the year, which affects the amount of solar radiation that reaches the ground, and therefore affects the risk of frozen ground.

9. Shading (SH): Represents the extent of shading on the ground, which affects the amount of solar radiation that reaches the ground, and therefore affects the risk of frozen ground.

10. Slope (SL): Represents the slope of the ground, which affects the rate of heat transfer between the ground and the air, and therefore affects the risk of frozen ground.

11. Aspect (AS): Represents the aspect of the ground, which affects the amount of solar radiation that reaches the ground, and therefore affects the risk of frozen ground.

12. Vegetation (VG): Represents the extent of vegetation cover, which affects the moisture content of the soil and the rate of heat transfer between the ground and the air, and therefore affects the risk of frozen ground.

13. Soil Type (ST): Represents the type of soil, which affects the thermal properties of the ground and the rate of heat transfer between the ground and the air, and therefore affects the risk of frozen ground.

14. Weather Station (WSN): Represents the location of a weather station, which provides measurements of the weather variables (AT, WS, RH, CC, TOD, DOY, SH, SL, AS, VG, and ST) that affect the risk of frozen ground.

The nodes in the Bayesian Network are connected by directed arrows that represent conditional dependencies between the variables. The direction of the arrows indicates the direction of causality between the variables. For example, the arrow pointing from AT to GT indicates that air temperature affects ground temperature. The network structure allows for the propagation of uncertainty between the variables, enabling the estimation of the risk of frozen ground based on the input variables.
\textbf{}

\section{Detailed statistics of the experiments}
\label{sec:app_detailed_experiment_stat}

\begin{figure*}[t!]
    \centering
    \includegraphics[scale=0.6]{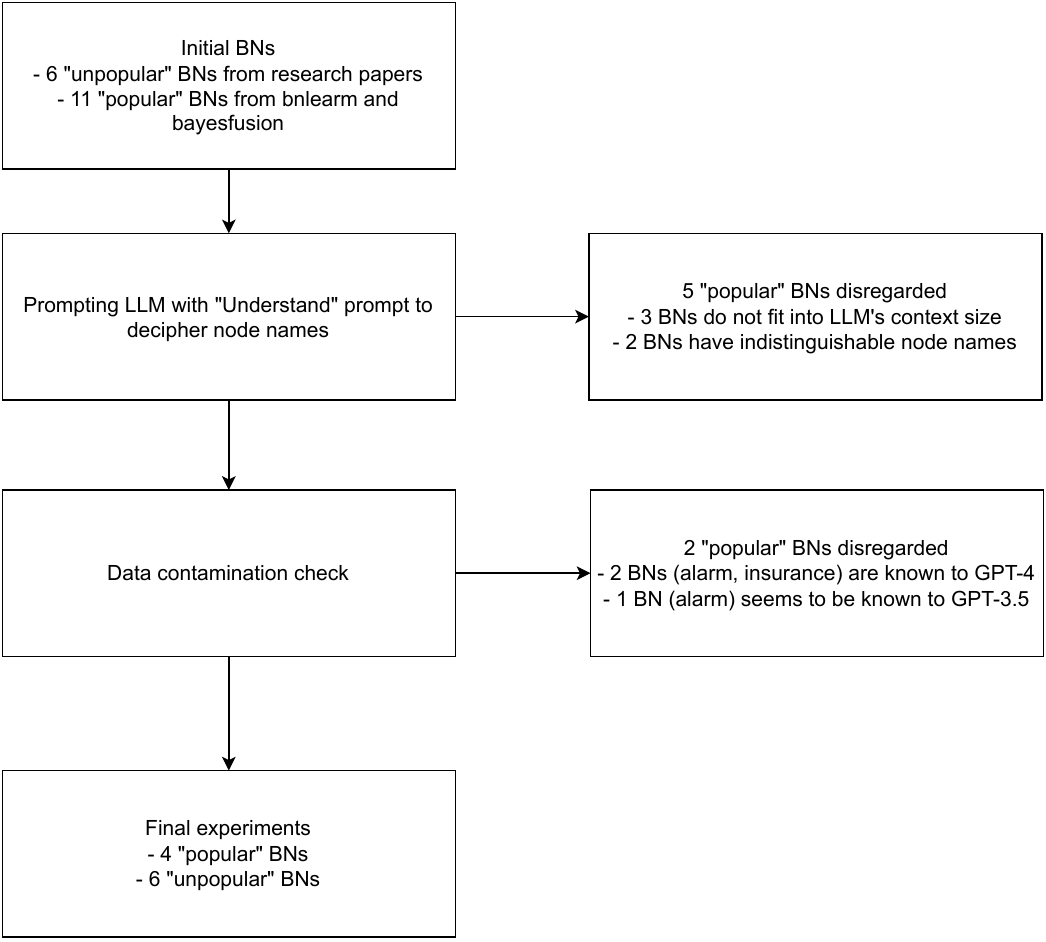}
    \caption{The process of BNs selection for final experiments.}
    \label{fig:bns_selection}
\end{figure*}

\begin{figure*}[t!]
    \centering
    \includegraphics[scale=0.6]{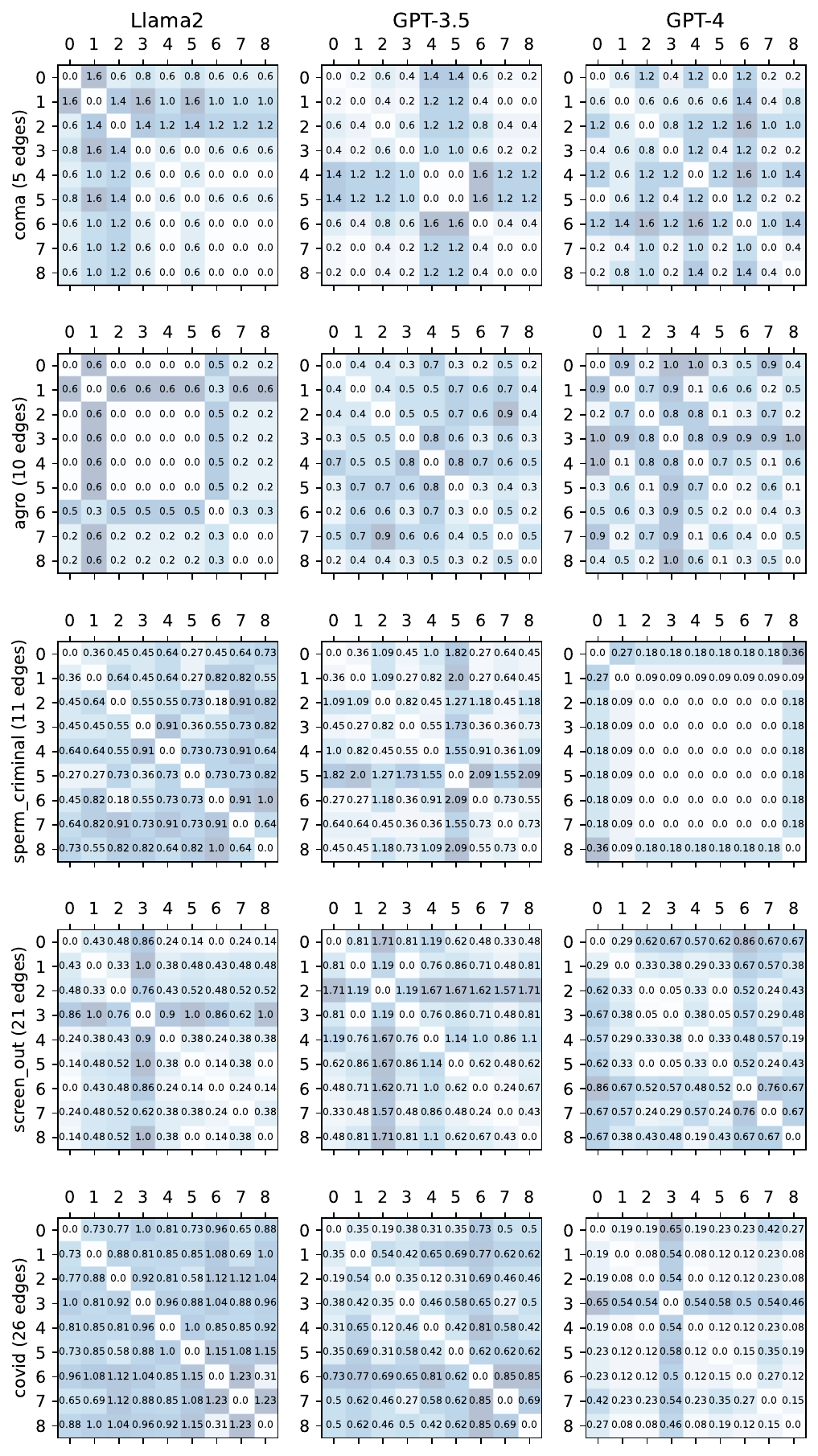}
    \caption{SHD between structures generated by `LLM-experts'' of all types for all BNs engaged in the study. Plot 1}
    \label{fig:experts_structure_diff_all_part1}
\end{figure*}

\begin{figure*}[t!]
    \centering
    \includegraphics[scale=0.6]{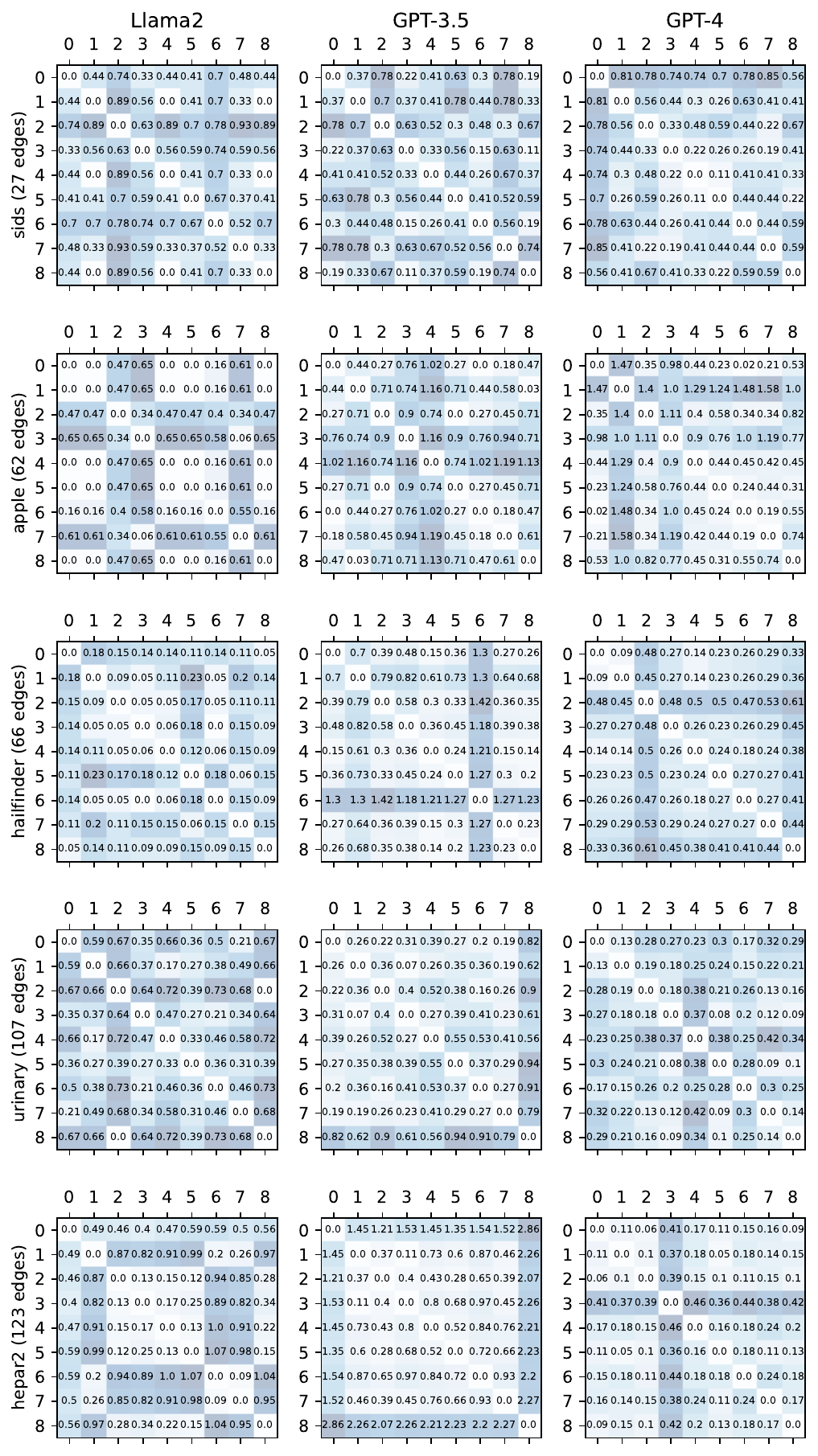}
    \caption{SHD between structures generated by `LLM-experts'' of all types for all BNs engaged in the study. Plot 2.}
    \label{fig:experts_structure_diff_all_part2}
\end{figure*}

\begin{figure*}[t!]
    \centering
    \includegraphics[scale=0.6]{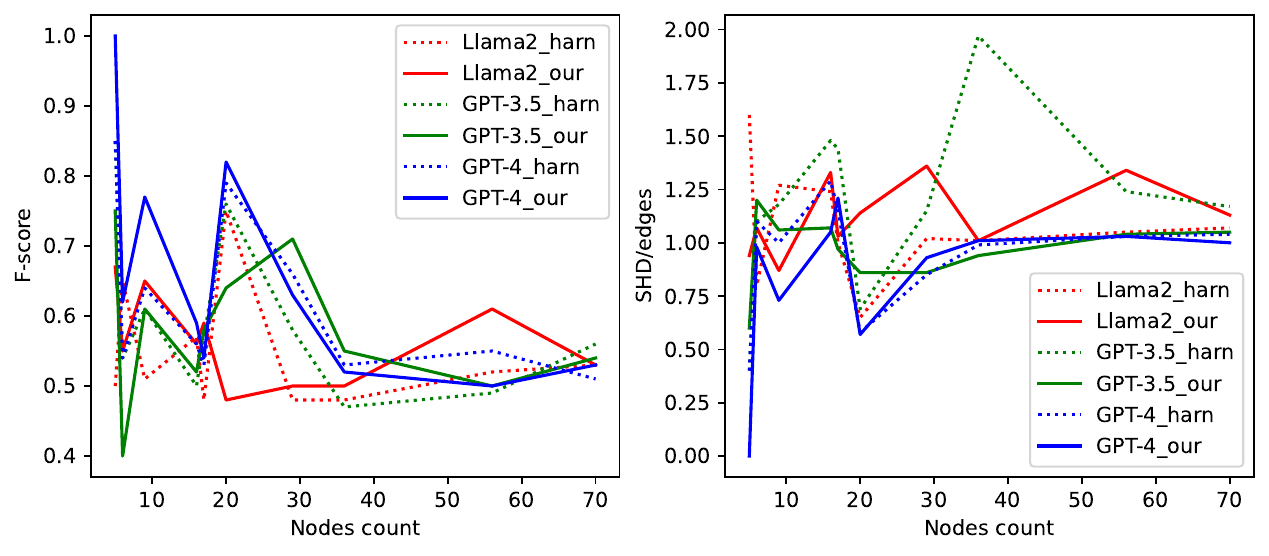}
    \caption{F-score and SHD resulting from both studied method with all LLMs related to the number of nodes in a BN.}
    \label{fig:fscore_shd_by_nodes}
\end{figure*}

\begin{table*}[]
\centering
\begin{tabular}{|c|c|cccc|cccc|}
\hline
\multirow{2}{*}{\textbf{BN}} & \multirow{2}{*}{\textbf{Method}} & \multicolumn{4}{c|}{\textbf{GPT-3.5}} & \multicolumn{4}{c|}{\textbf{Llama2}} \\ \cline{3-10} 
 &  & \multicolumn{1}{c|}{\textbf{FP}} & \multicolumn{1}{c|}{\textbf{FN}} & \multicolumn{1}{c|}{\textbf{F-score}} & \textbf{SHD/edg} & \multicolumn{1}{c|}{\textbf{FP}} & \multicolumn{1}{c|}{\textbf{FN}} & \multicolumn{1}{c|}{\textbf{F-score}} & \textbf{SHD/edg} \\ \hline
\multirow{2}{*}{coma} & har & \multicolumn{1}{c|}{0} & \multicolumn{1}{c|}{2} & \multicolumn{1}{c|}{0.85} & 0.4 & \multicolumn{1}{c|}{4} & \multicolumn{1}{c|}{4} & \multicolumn{1}{c|}{0.5} & 1.6 \\ \cline{2-10} 
 & our & \multicolumn{1}{c|}{0} & \multicolumn{1}{c|}{3} & \multicolumn{1}{c|}{0.75} & 0.6 & \multicolumn{1}{c|}{2} & \multicolumn{1}{c|}{3} & \multicolumn{1}{c|}{0.67} & 0.94 \\ \hline
\multirow{2}{*}{agro} & har & \multicolumn{1}{c|}{3} & \multicolumn{1}{c|}{8} & \multicolumn{1}{c|}{0.54} & 1.1 & \multicolumn{1}{c|}{1} & \multicolumn{1}{c|}{7} & \multicolumn{1}{c|}{0.65} & 0.8 \\ \cline{2-10} 
 & our & \multicolumn{1}{c|}{2} & \multicolumn{1}{c|}{10} & \multicolumn{1}{c|}{0.4} & 1.2 & \multicolumn{1}{c|}{3} & \multicolumn{1}{c|}{8} & \multicolumn{1}{c|}{0.55} & 1.07 \\ \hline
\multirow{2}{*}{sperm} & har & \multicolumn{1}{c|}{5} & \multicolumn{1}{c|}{8} & \multicolumn{1}{c|}{0.61} & 1.18 & \multicolumn{1}{c|}{4} & \multicolumn{1}{c|}{10} & \multicolumn{1}{c|}{0.51} & 1.27 \\ \cline{2-10} 
 & our & \multicolumn{1}{c|}{3} & \multicolumn{1}{c|}{8} & \multicolumn{1}{c|}{0.61} & 1.06 & \multicolumn{1}{c|}{1} & \multicolumn{1}{c|}{8} & \multicolumn{1}{c|}{0.65} & 0.87 \\ \hline
\multirow{2}{*}{screen} & har & \multicolumn{1}{c|}{11} & \multicolumn{1}{c|}{20} & \multicolumn{1}{c|}{0.5} & 1.48 & \multicolumn{1}{c|}{8} & \multicolumn{1}{c|}{18} & \multicolumn{1}{c|}{0.57} & 1.24 \\ \cline{2-10} 
 & our & \multicolumn{1}{c|}{3} & \multicolumn{1}{c|}{20} & \multicolumn{1}{c|}{0.52} & 1.07 & \multicolumn{1}{c|}{10} & \multicolumn{1}{c|}{18} & \multicolumn{1}{c|}{0.56} & 1.33 \\ \hline
\multirow{2}{*}{covid} & har & \multicolumn{1}{c|}{3} & \multicolumn{1}{c|}{15} & \multicolumn{1}{c|}{0.76} & 0.69 & \multicolumn{1}{c|}{0} & \multicolumn{1}{c|}{17} & \multicolumn{1}{c|}{0.75} & 0.65 \\ \cline{2-10} 
 & our & \multicolumn{1}{c|}{1} & \multicolumn{1}{c|}{21} & \multicolumn{1}{c|}{0.64} & 0.86 & \multicolumn{1}{c|}{4} & \multicolumn{1}{c|}{26} & \multicolumn{1}{c|}{0.48} & 1.14 \\ \hline
\multirow{2}{*}{sids} & har & \multicolumn{1}{c|}{17} & \multicolumn{1}{c|}{22} & \multicolumn{1}{c|}{0.57} & 1.44 & \multicolumn{1}{c|}{0} & \multicolumn{1}{c|}{27} & \multicolumn{1}{c|}{0.48} & 1 \\ \cline{2-10} 
 & our & \multicolumn{1}{c|}{3} & \multicolumn{1}{c|}{23} & \multicolumn{1}{c|}{0.58} & 0.97 & \multicolumn{1}{c|}{5} & \multicolumn{1}{c|}{23} & \multicolumn{1}{c|}{0.59} & 1.03 \\ \hline
\multirow{2}{*}{apple} & har & \multicolumn{1}{c|}{18} & \multicolumn{1}{c|}{53} & \multicolumn{1}{c|}{0.58} & 1.15 & \multicolumn{1}{c|}{1} & \multicolumn{1}{c|}{62} & \multicolumn{1}{c|}{0.48} & 1.02 \\ \cline{2-10} 
 & our & \multicolumn{1}{c|}{13} & \multicolumn{1}{c|}{40} & \multicolumn{1}{c|}{0.71} & 0.86 & \multicolumn{1}{c|}{24} & \multicolumn{1}{c|}{60} & \multicolumn{1}{c|}{0.5} & 1.36 \\ \hline
\multirow{2}{*}{hailfinder} & har & \multicolumn{1}{c|}{70} & \multicolumn{1}{c|}{66} & \multicolumn{1}{c|}{0.49} & 1.24 & \multicolumn{1}{c|}{5} & \multicolumn{1}{c|}{64} & \multicolumn{1}{c|}{0.52} & 1.05 \\ \cline{2-10} 
 & our & \multicolumn{1}{c|}{3} & \multicolumn{1}{c|}{65} & \multicolumn{1}{c|}{0.5} & 1.04 & \multicolumn{1}{c|}{37} & \multicolumn{1}{c|}{52} & \multicolumn{1}{c|}{0.61} & 1.34 \\ \hline
\multirow{2}{*}{urinary} & har & \multicolumn{1}{c|}{108} & \multicolumn{1}{c|}{103} & \multicolumn{1}{c|}{0.47} & 1.97 & \multicolumn{1}{c|}{1} & \multicolumn{1}{c|}{107} & \multicolumn{1}{c|}{0.48} & 1.01 \\ \cline{2-10} 
 & our & \multicolumn{1}{c|}{2} & \multicolumn{1}{c|}{99} & \multicolumn{1}{c|}{0.55} & 0.94 & \multicolumn{1}{c|}{3} & \multicolumn{1}{c|}{105} & \multicolumn{1}{c|}{0.5} & 1.01 \\ \hline
\multirow{2}{*}{hepar2} & har & \multicolumn{1}{c|}{36} & \multicolumn{1}{c|}{112} & \multicolumn{1}{c|}{0.56} & 1.17 & \multicolumn{1}{c|}{15} & \multicolumn{1}{c|}{117} & \multicolumn{1}{c|}{0.53} & 1.07 \\ \cline{2-10} 
 & our & \multicolumn{1}{c|}{13} & \multicolumn{1}{c|}{117} & \multicolumn{1}{c|}{0.54} & 1.05 & \multicolumn{1}{c|}{23} & \multicolumn{1}{c|}{116} & \multicolumn{1}{c|}{0.53} & 1.13 \\ \hline
\end{tabular}
\caption{Detailed information about the performance of \textit{harness} and our method with GPT-3.5 and Llama2. FP and FN correspond to false positive and false negative edges in the learned BN structure respectively.}
\label{tab:gpt35_and_llama2_expanded_metrics}
\end{table*}

\FloatBarrier

\section{Error analysis}
\label{sec:appendix_error_analysis}

\textbf{Llama2 generates BN structure using non-existing node obtained from its system message}

\textit{System message}

You will generate a predictive model using a specialized communication protocol.  Assume the presence of multiple specialists possessing the required skills in the designated problem domain. Each specialist responds independently to our questions. Provide input as an expert with the following profile: Epidemiologist with 15 years of experience in studying the distribution and determinants of health-related events, diseases, or health-related characteristics among populations, including the epidemiology of \textbf{urinary tract infections} in children.

\textit{User prompt}

Consider the factors associated with the predictive model, represented by the list of nodes < Age group >,< Sex >,< Comorbidity >,< Diarrhoea >,< On antibiotics upon presentation to ED >,< level of colonisation of the perineum/ external genitalia by E. coli >,< level of colonisation of the perineum/ external genitalia by non-E. coli gram negative bacteria >,< level of colonisation of the perineum/ external genitalia by gram positive bacteria >,< Pathogenic organism infecting the urinary tract and causing disease >,< Current clinical phenotype >,< Empiric antibiotic prescription received during the ED presentation >,< Urine collection method >,< Assessment of epithelial cells via microscopic or automated analysis >,< risk of a non-causative organism/s entering the urine specimen during the specimen collection process >,< Assessment of bacteria via microscopic analysis >,< isolation of E.coli from the urine specimen >,< isolation of non-E.coli gram negative bacteria from the urine specimen >,< isolation of gram positive bacteria from the urine specimen >,< leukocyte esterase detected on urine dipstick >,< Nitrite detected on the urine dipstick >,< assessment of leucocytes level in the urine specimen via microscopic analysis >,< Leukocyte counts in blood >,< C-Reactive protein in blood >,< Absolute neutrophil counts in blood >,< Genital pain (or discomfort), dysuria, and urinary tract relevant discomfortn >,< Abdominal pain >,< Haematuria >,< Foul smelling urine >,< Result of ultrasound investigation >,< Respiratory symptoms >,< Parent reported fever in medical notes >,< Temperature (degrees celsius) >,< Irritability >,< Lethargy >,< Nausea and or vomiting >,< Poor oral intake >.      Now, analyze the relationships between these factors. There are three possible types of relations:              Factor A directly affects Factor B       Factor B directly affects Factor A       No direct effect between two factors

\textit{System reply}

....

\textit{User prompt}

.....

\textit{System reply}

Here is a valid JSON representation of the relationships between the specified factors:```json[ ["Age group", "\textbf{Urinary tract infection}"], ["Sex", "\textbf{Urinary tract infection}"], ["Comorbidity", "\textbf{Urinary tract infection}"], ["Diarrhea", "\textbf{Urinary tract infection}"], ["On antibiotics upon presentation to ED", "\textbf{Urinary tract infection}"], ["Level of colonisation of the perineum/external genitalia by E. coli", "\textbf{Urinary tract infection}"], ["Level of colonisation of the perineum/external genitalia by non-E. coli gram negative bacteria", "\textbf{Urinary tract infection}"], ["Level of colonisation of the perineum/external genitalia by gram positive bacteria", "\textbf{Urinary tract infection}"], ["Pathogenic organism infecting the urinary tract and causing disease", "\textbf{Urinary tract infection}"], ["Current clinical phenotype", "\textbf{Urinary tract infection}"], ["Empiric antibiotic prescription received during the ED presentation", "\textbf{Urinary tract infection}"], ["Urine collection method", "\textbf{Urinary tract infection}"], ["Assessment of epithelial cells via microscopic or automated analysis", "\textbf{Urinary tract infection}"], ["Risk of a non-causative organism/s entering the urine specimen during the specimen collection process", "\textbf{Urinary tract infection}"], ["Assessment of bacteria via microscopic analysis", "\textbf{Urinary tract infection}"], ["Isolation of E.coli from the urine specimen", "\textbf{Urinary tract infection}"], ["Isolation of non-E.coli gram negative bacteria from the urine specimen", "\textbf{Urinary tract infection}"], ["Isolation of gram positive bacteria from the urine specimen", "\textbf{Urinary tract infection}"], ["Leukocyte esterase detected on urine dipstick", "\textbf{Urinary tract infection}"], ["Nitrite detected on the urine dipstick", "\textbf{Urinary tract infection}"], ["Assessment of leucocytes level in the urine specimen via microscopic analysis", "\textbf{Urinary tract infection}"], ["Leukocyte counts in blood", "\textbf{Urinary tract infection}"], ["C-Reactive protein in blood", "\textbf{Urinary tract infection}"], ["Absolute neutrophil counts in blood", "\textbf{Urinary tract infection}"], ["Genital pain (or discomfort), dysuria, and urinary tract relevant discomfort", "\textbf{Urinary tract infection}"], ["Abdominal pain", "\textbf{Urinary tract infection}"], ["Haematuria", "\textbf{Urinary tract infection}"], ["Foul smelling urine", "\textbf{Urinary tract infection}"], ["Result of ultrasound investigation", "\textbf{Urinary tract infection}"], ["Respiratory symptoms", "\textbf{Urinary tract infection}"], ["Parent reported fever in medical notes", "\textbf{Urinary tract infection}"], ["Temperature (degrees celsius)", "\textbf{Urinary tract infection}"], ["Irritability", "\textbf{Urinary tract infection}"], ["Lethargy", "\textbf{Urinary tract infection}"], ["Nausea and or vomiting", "\textbf{Urinary tract infection}"], ["Poor oral intake", "\textbf{Urinary tract infection}"]]```Note that this representation only includes the direct relationships between the factors and does not include any indirect relationships or interactions between factors.

\begin{table}[]
\small
\centering
\begin{tabular}{|l|l|l|l|}
\hline
\textbf{Method} & \textbf{Llama2} & \textbf{GPT-3.5} & \textbf{GPT-4} \\ \hline
Harness & 0 & 1.2 & 0 \\ \hline
our & 0.6 & 15.6 & 0.6 \\ \hline
\end{tabular}
\caption{Average number of cycles generated by both methods with all LLMs for all BNs}
\label{tab:cycles_average}
\end{table}

\begin{table}[]
\centering
\begin{tabular}{|l|l|l|l|}
\hline
\textbf{BN} & \textbf{Llama2} & \textbf{GPT-3.5} & \textbf{GPT-4} \\ \hline
coma & 3 & 6 & 0 \\ \hline
agro & 0 & 11 & 2 \\ \hline
sperm & 3 & 49 & 0 \\ \hline
screen & 0 & 60 & 1 \\ \hline
covid & 0 & 2 & 0 \\ \hline
sids & 0 & 23 & 0 \\ \hline
apple & 0 & 4 & 2 \\ \hline
hailfinder & 0 & 0 & 0 \\ \hline
urinary & 0 & 0 & 0 \\ \hline
hepar2 & 0 & 1 & 1 \\ \hline
\end{tabular}
\caption{Cycles generated by LLMs for different BNs when using our method}
\label{tab:duplicates_expanded}
\end{table}

\textbf{GPT-4 refuses to generate a JSON with urinary BN structure}


\textit{System reply (to the second prompt of our method)}

As a medical ethicist, I am not qualified to determine the direct relationships between these medical factors. My role is to ensure that the process of building this predictive model respects the ethical principles of autonomy, beneficence, non-maleficence, and justice. I recommend consulting with a medical professional or a data scientist to determine these relationships.

\textbf{Llama2 generates a ``cause-and-effect chain'' rather than BN structure (hailfinder BN)}

\textit{System message}

You will generate a predictive model using a specialized communication protocol.    Assume the presence of multiple specialists possessing the required skills in the designated problem domain.    Each specialist responds independently to our questions. Provide input as an expert with the following profile: Communication and collaboration expert with a background in psychology, sociology, or organizational behavior, skilled in facilitating effective communication and collaboration among diverse groups

\textit{User prompt}

Consider the factors associated with the predictive model, represented by the list of nodes < relationship between the scenario and the morning convective inhibition >,< level of morning convective inhibition >,< morning convective inhibition in the given scenario >,< atmospheric moisture conditions >,< relationship between the scenario and the morning atmospheric instability >,< Lifted Index (LI) and density relationship >,< morning atmospheric instability in the given scenario >,< morning atmospheric instability >,< combined vertical motion >,< mesoscale area >,< combined moisture conditions >,< area of dry air >,< outflow from the mountains >,< wind hodograph >,< morning boundaries >,< overall boundaries >,<  Composite Planetary Forecast >,< Convective Available Potential Energy (CAPE) Change >,< Convective Available Potential Energy (CAPE) in Scenario >,<  Instability in Mountainous Terrain >,< Cloud Shade due to Convection >,< Combined Clouds >,< Other Cloud Shade >,< Visible Cloud Cover >,< Infrared Cloud Cover >,< Saturation Content Moisture >,< Rao Content Moisture >,< 0.7 Micrometer Vertical Motion >,< Subjective Vertical Motion >,<  Quasi-Geostrophic Vertical Motion >,<  Latest Convective Available Potential Energy (CAPE) Inhibition >,< Lifted Level Index Weighted >,<  Current Propagation of Convection >,< Date of Weather Forecast >,< Weather Scenario >,< Atmospheric Dewpoints >,<  Low-Level Moisture Advection >,< Instability Change >,< Instability Scale in Scenario >,< Low-Level Lapse Rate >,< Mean Relative Humidity >,< Mid-Level Lapse Rate >,< Mountainous Forecast >,< Movement Features >,< Scenario Relative to 3-4 >,< Plains Forecast >,< North 34-Star Forecast >,< Scenario Relative Planetary Forecast >,< Region 5 Forecast >,< Relative Humidity Ratio >,< Surface Wind Shift Distance >,< Synoptic Forcing >,< Temperature Distribution >,< Wind Aloft >,< Wind Field in Mountainous Terrain >,< Wind Field in Plains >.      Now, analyze the relationships between these factors. There are three possible types of relations:              Factor A directly affects Factor B       Factor B directly affects Factor A       No direct effect between two factors              Please systematically evaluate the interconnections between the specified factors, focusing only on significant relation

\textit{System reply}

As a communication and collaboration expert with a background in psychology, sociology, and organizational behavior, I will facilitate effective communication and collaboration among diverse groups to analyze the relationships between the factors associated with the predictive model. We will evaluate the interconnections between the specified factors, focusing only on significant relations.1. Relationship between the scenario and the morning convective inhibition:\t* Factor A (scenario) directly affects Factor B (morning convective inhibition)2. Level of morning convective inhibition:\t* Factor A (morning convective inhibition) directly affects Factor B (atmospheric moisture conditions)3. Morning convective inhibition in the given scenario:\t* Factor A (scenario) directly affects Factor B (morning convective inhibition)4. Atmospheric moisture conditions:\t* Factor A (atmospheric moisture conditions) directly affects Factor B (morning convective inhibition)5. Relationship between the scenario and the morning atmospheric instability:\t* Factor A (scenario) directly affects Factor B (morning atmospheric instability)6. Lifted Index (LI) and density relationship:* Factor A (Lifted Index (LI)) directly affects Factor B (density relationship)7. Morning atmospheric instability in the given scenario:* Factor A (scenario) directly affects Factor B (morning atmospheric instability)8. Morning atmospheric instability:* Factor A (morning atmospheric instability) directly affects Factor B (combined vertical motion)9. Combined vertical motion:* Factor A (combined vertical motion) directly affects Factor B (mesoscale area)10. Mesoscale area:* Factor A (mesoscale area) directly affects Factor B (combined moisture conditions)11. Combined moisture conditions:* Factor A (combined moisture conditions) directly affects Factor B (area of dry air)12. Area of dry air:* Factor A (area of dry air) directly affects Factor B (outflow from the mountains)13. Outflow from the mountains:* Factor A (outflow from the mountains) directly affects Factor B (wind hodograph)14. Wind hodograph:* Factor A (wind hodograph) directly affects Factor B (morning boundaries)15. Morning boundaries:* Factor A (morning boundaries) directly affects Factor B (overall boundaries)16. Overall boundaries:* Factor A (overall boundaries) directly affects Factor B (Composite Planetary Forecast)17. Composite Planetary Forecast:* Factor A (Composite Planetary Forecast) directly affects Factor B (Convective Available Potential Energy (CAPE) Change)18. Convective Available Potential Energy (CAPE) Change:* Factor A (Convective Available Potential Energy (CAPE) Change) directly affects Factor B (Convective Available Potential Energy (CAPE) in Scenario)19. Convective Available Potential Energy (CAPE) in Scenario:* Factor A (Convective Available Potential Energy (CAPE) in Scenario) directly affects Factor B (Instability in Mountainous Terrain)20. Instability in Mountainous Terrain:* Factor A (Instability in Mountainous Terrain) directly affects Factor B (Cloud Shade due to Convection)21. Cloud Shade due to Convection:* Factor A (Cloud Shade due to Convection) directly affects Factor B (Combined Clouds)22. Combined Clouds:* Factor A (Combined Clouds) directly affects Factor B (Other Cloud Shade)23. Other Cloud Shade:* Factor A (Other Cloud Shade) directly affects Factor B (Visible Cloud Cover)24. Visible Cloud Cover:* Factor A (Visible Cloud Cover) directly affects Factor B (Infrared Cloud Cover)25. Infrared Cloud Cover:* Factor A (Infrared Cloud Cover) directly affects Factor B (Saturation Content Moisture)26. Saturation Content Moisture:* Factor A (Saturation Content Moisture) directly affects Factor B (Rao Content Moisture)27. Rao Content Moisture:* Factor A (Rao Content Moisture) directly affects Factor B (0.7 Micrometer Vertical Motion)28.0.7 Micrometer Vertical Motion:* Factor A (0.7 Micrometer Vertical Motion)

\textit{User prompt}

....

\textit{System reply}

Here is a valid JSON representation of the relationships between the specified factors:```json[ ["weather Scenario", "level of Morning Convective Inhibition"], ["level of Morning Convective Inhibition", "Atmospheric Moisture Conditions"], ["Atmospheric Moisture Conditions", "Lifted Index (LI) and Density Relationship"], ["Lifted Index (LI) and Density Relationship", "Morning Atmospheric Instability"], ["Morning Atmospheric Instability", "Combined Vertical Motion"], ["Combined Vertical Motion", "Mesoscale Area"], ["Mesoscale Area", "Combined Moisture Conditions"], ["Combined Moisture Conditions", "Area of Dry Air"], ["Area of Dry Air", "Outflow from the Mountains"], ["Outflow from the Mountains", "Wind Hodograph"], ["Wind Hodograph", "Morning Boundaries"], ["Morning Boundaries", "Overall Boundaries"], ["Overall Boundaries", "Composite Planetary Forecast"], ["Composite Planetary Forecast", "Convective Available Potential Energy (CAPE) Change"], ["Convective Available Potential Energy (CAPE) Change", "Convective Available Potential Energy (CAPE) in Scenario"], ["Convective Available Potential Energy (CAPE) in Scenario", "Instability in Mountainous Terrain"], ["Instability in Mountainous Terrain", "Cloud Shade due to Convection"], ["Cloud Shade due to Convection", "Combined Clouds"], ["Combined Clouds", "Other Cloud Shade"], ["Other Cloud Shade", "Visible Cloud Cover"], ["Visible Cloud Cover", "Infrared Cloud Cover"], ["Infrared Cloud Cover", "Saturation Content Moisture"], ["Saturation Content Moisture", "Rao Content Moisture"], ["Rao Content Moisture", "0.7 Micrometer Vertical Motion"], ["0.7 Micrometer Vertical Motion", "Subjective Vertical Motion"], ["Subjective Vertical Motion", "Quasi-Geostrophic Vertical Motion"], ["Quasi-Geostrophic Vertical Motion", "Latest Convective Available Potential Energy (CAPE) Inhibition"], ["Latest Convective Available Potential Energy (CAPE) Inhibition", "Lifted Level Index Weighted"], ["Lifted Level Index Weighted", "Current Propagation of Convection"], ["Current Propagation of Convection", "Date of Weather Forecast"], ["Date of Weather Forecast", "Weather Scenario"], ["Weather Scenario", "Atmospheric Dewpoints"], ["Atmospheric Dewpoints", "Low-Level Moisture Advection"], ["Low-Level Moisture Advection", "Instability Change"], ["Instability Change", "Instability Scale in Scenario"], ["Instability Scale in Scenario", "Low-Level Lapse Rate"], ["Low-Level Lapse Rate", "Mean Relative Humidity"], ["Mean Relative Humidity", "Mid-Level Lapse Rate"], ["Mid-Level Lapse Rate", "Mountainous Forecast"], ["Mountainous Forecast", "Movement Features"], ["Movement Features", "Scenario Relative to 3-4"], ["Scenario Relative to 3-4", "Plains Forecast"], ["Plains Forecast", "North 34-Star Forecast"], ["North 34-Star Forecast", "Scenario Relative Planetary Forecast"], ["Scenario Relative Planetary Forecast", "Region 5 Forecast"], ["Region 5 Forecast", "Relative Humidity Ratio"], ["Relative Humidity Ratio", "Surface Wind Shift Distance"], ["Surface Wind Shift Distance", "Synoptic Forcing"], ["Synoptic Forcing", "Temperature Distribution"], ["Temperature Distribution", "Wind Aloft"], ["Wind Aloft", "Wind Field in Mountainous Terrain"], ["Wind Field in Mountainous Terrain", "Wind Field in Plains"]]```

\section{Future work}
\label{sec:future_work}

There are several ways to improve the performance of the described algorithms. For both \textit{Harness} and our method, one may try to apply the algorithm for the decreased number of nodes several times, until all possible pairs of nodes have been located inside one request at least once. For our method, more interaction between different ``LLM experts'' could be established. Similarly to the practices used in~\cite{xiong2023examining,zhuge2023mindstorms} it may be useful to engage different experts in a discussion of their original request, and then independently ask whether they want to change their original decision after the discussion. Finally, it may be interesting to fine-tune LLMs to the specific task of BNSL using the existing BNs.

\end{document}